%% file: template.tex
\documentclass[journal, hideappendix]{vgtc}                     % final (journal style)
\newcommand{\eg}{\textit{e.g.}}

\newcommand{\etal}{\textit{et al.}}

\newcommand{\red}[1]{{\color{black}{#1}}}

\onlineid{1292}

\vgtccategory{Representations \& Interaction}

\title{Let the Chart Spark: Embedding Semantic Context into Chart with Text-to-Image Generative Model}

\author{%
  Shishi Xiao,
  Suizi Huang,
  Yue Lin,
  Yilin Ye,
  Wei Zeng
}

\authorfooter{
   \item
   Shishi Xiao, Suizi Huang, Yue Lin, Yilin Ye, and Wei Zeng are with the Hong Kong University of Science and Technology (Guangzhou), Guangzhou, China. Yilin Ye and Wei Zeng are also with the Hong Kong University of Science and Technology, Hong Kong SAR, China.
   E-mail: \{sxiao713@connect., shuang310@connect., ylin491@connect., yyebd@connect., weizeng@\}hkust-gz.edu.cn
   \item
   Wei Zeng is the corresponding author.
}
\input{Latex/0_Abstract}


\teaser{
  \centering
  \includegraphics[width=\linewidth]{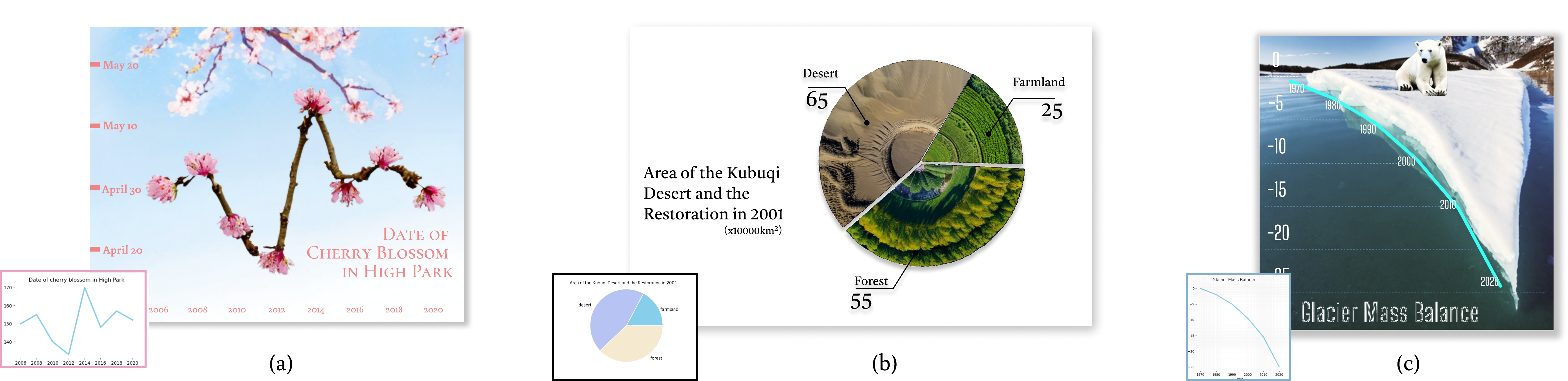}
  \caption{\textbf{Pictorial visualizations created by \textit{ChartSpark}.} (a) a line chart depicts the date of cherry blossom in High Park each year, embedded with the tree branch while also preserving the trend. (b) A pie chart shows the area of Kubuqi desert and its restoration in 2021, with the three types of land embedded in the corresponding sectors in a consistent style. (c) A line chart shows the amount of glacier mass balance per year, coherently presented with the background glacier structure in compliance with the data trend.
  }
  \label{fig:teaser}
}

\graphicspath{{figs/}{figures/}{pictures/}{images/}{./}} % where to search for the images

\usepackage{tabu}                      % only used for the table example
\usepackage{booktabs}                  % only used for the table example
\usepackage{lipsum}                    % used to generate placeholder text
\usepackage{mwe}                       % used to generate placeholder figures

\usepackage{mathptmx}                  % use matching math font
\usepackage{wrapfig}
\usepackage{graphicx}

\begin{document}

%%%%%%%%%%%%%%%%%%%%%%%%%%%%%%%%%%%%%%%%%%%%%%%%%%%%%%%%%%%%%%%%
%%%%%%%%%%%%%%%%%%%%%% START OF THE PAPER %%%%%%%%%%%%%%%%%%%%%%
%%%%%%%%%%%%%%%%%%%%%%%%%%%%%%%%%%%%%%%%%%%%%%%%%%%%%%%%%%%%%%%%


\maketitle
\input{Latex/0_Abstract}
\input{Latex/1_Introduction}
\input{Latex/2_RelatedWorks}

\input{Latex/3_Preliminary}
\input{Latex/4_Method}
\input{Latex/5_Interface}
\input{Latex/6_Evaluation}

\input{Latex/7_Discussion}

\input{Latex/8_Conclusion}

% The authors wish to thank anonymous reviewers for their constructive comments. This work was supported in part by the National Natural Science Foundation of China (No. 62172398).}

\bibliographystyle{abbrv-doi}
\bibliography{template}

\clearpage

% % \section*{Appendix}

% \includepdf[pages=-]{figs/Supplementary_final.pdf}

\begin{figure*}[t]
    \centering
    \includegraphics[width=0.9\linewidth]{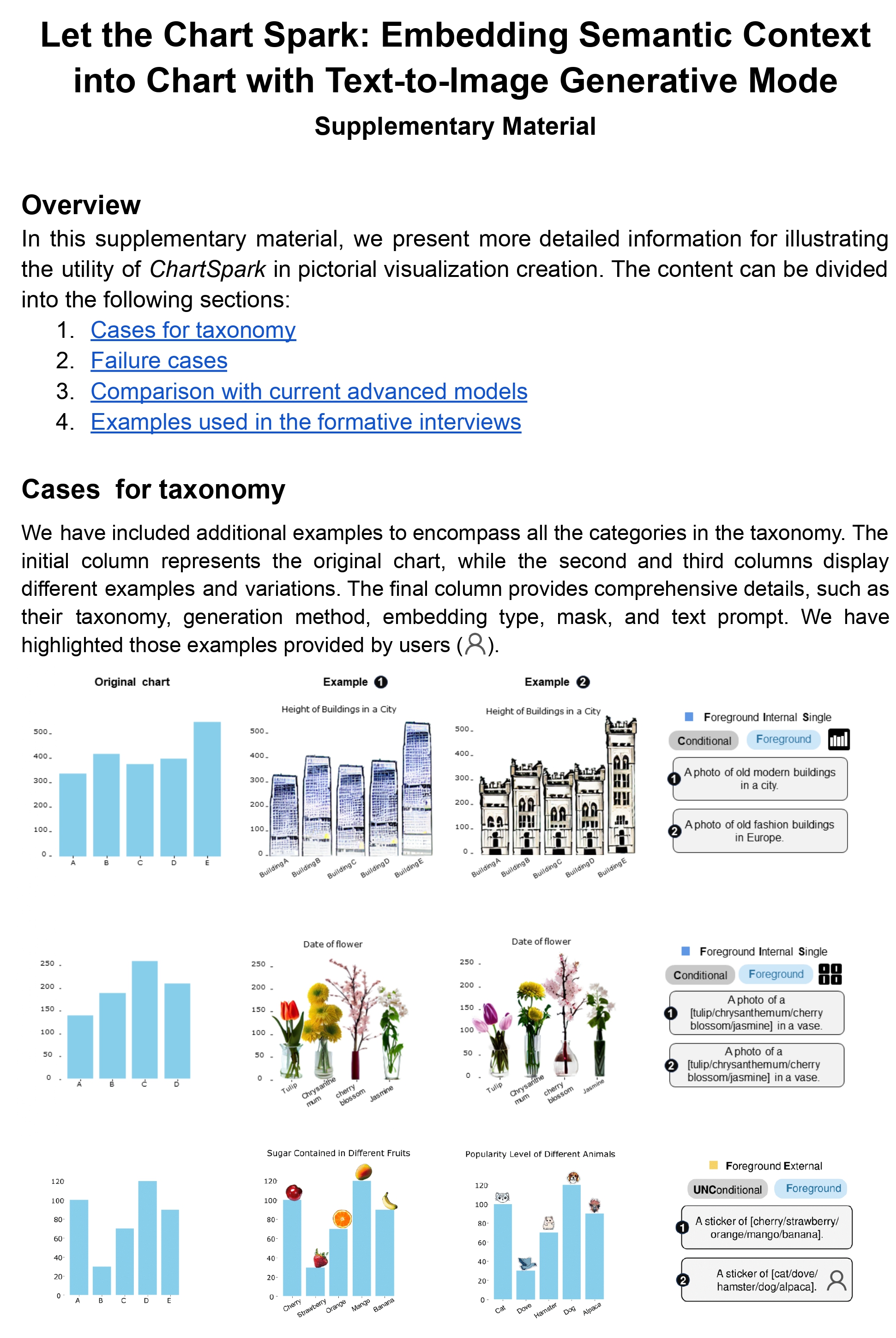}
    \vspace{6em}
    \caption{}
\end{figure*}

\begin{figure*}[t]
    \centering
    \vspace{-1em}
    \includegraphics[width=0.86\linewidth]{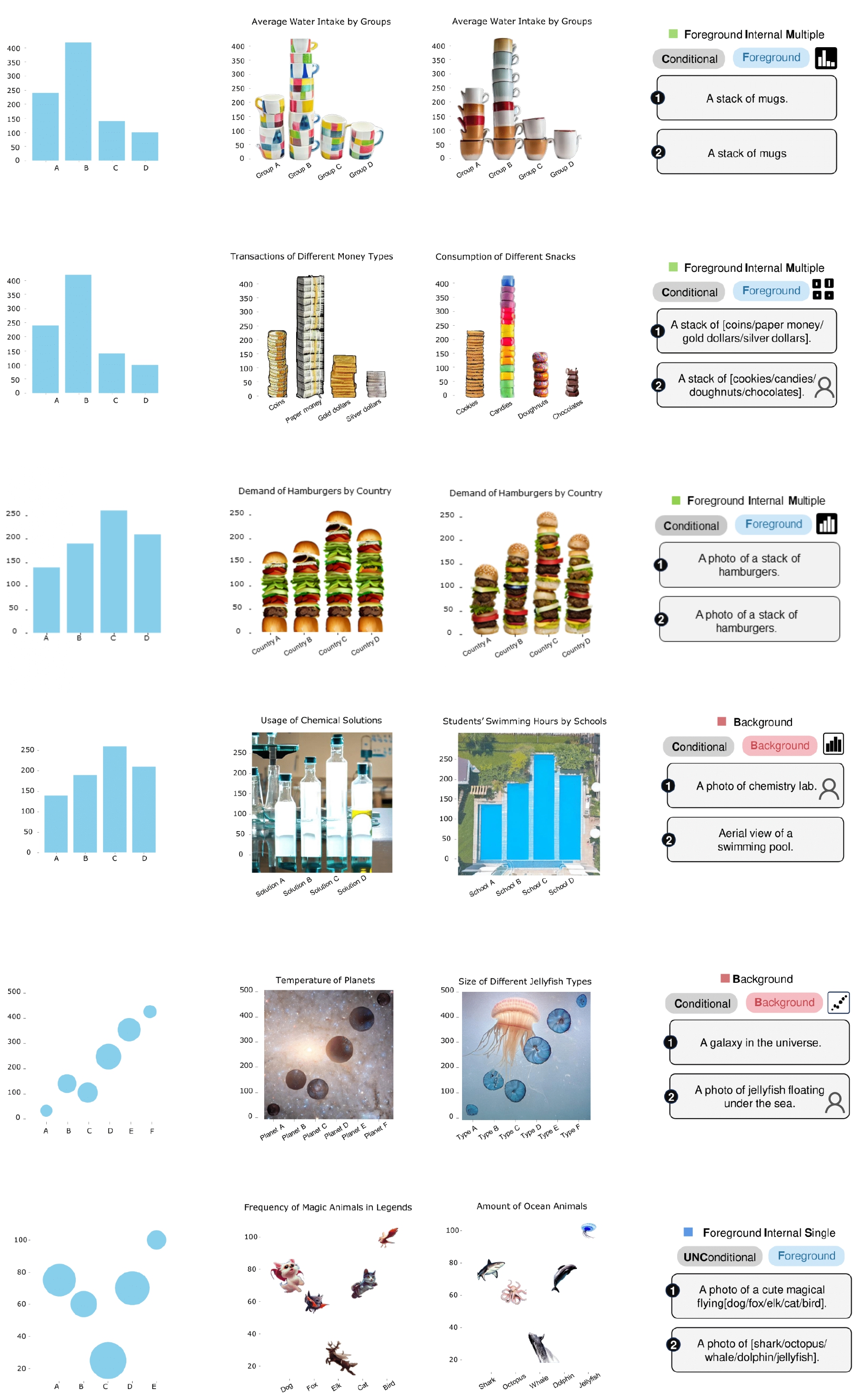}
    \vspace{6em}
    \caption{}
\end{figure*}

\begin{figure*}[t]
    \centering
    \vspace{-1.5em}
    \hspace{1.5em}
    \includegraphics[width=0.86\linewidth]{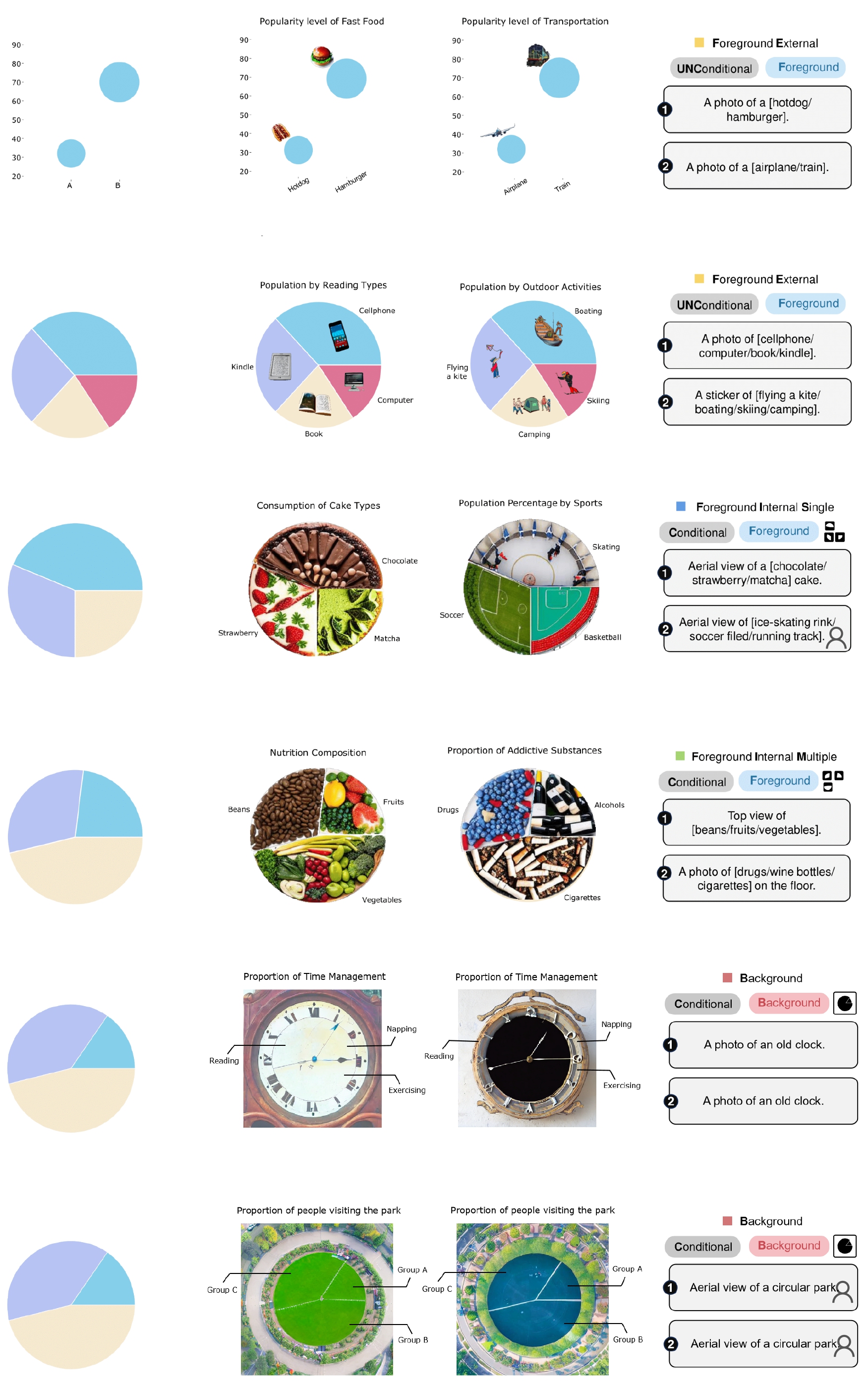}
    \vspace{6em}
    \caption{}
    
\end{figure*}
\begin{figure*}[t]
    \centering
    \includegraphics[width=0.9\linewidth]{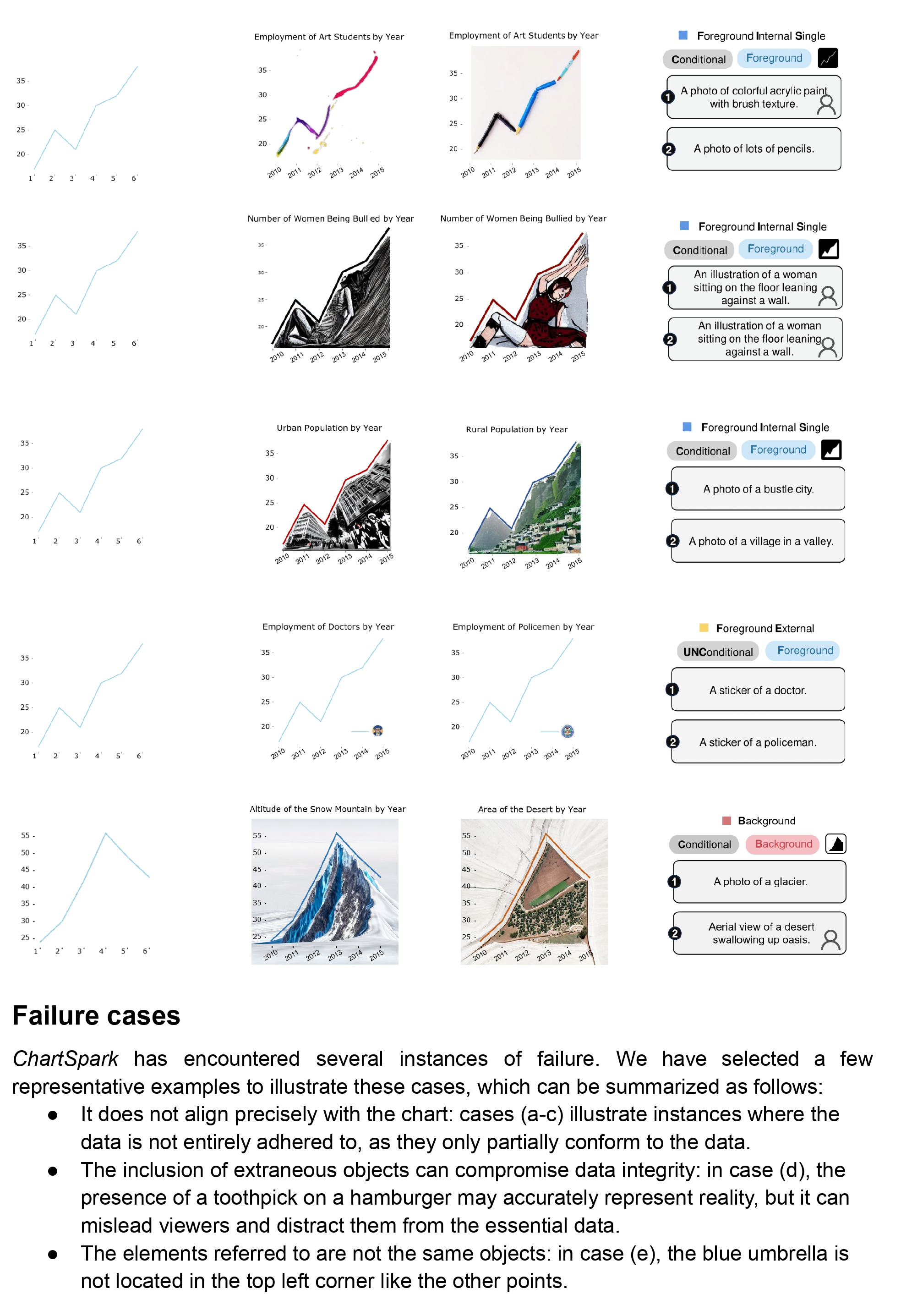}
    \vspace{8em}
    \caption{}
\end{figure*}

\begin{figure*}[t]
    \centering
    \vspace{-1em}
    \includegraphics[width=0.9\linewidth]{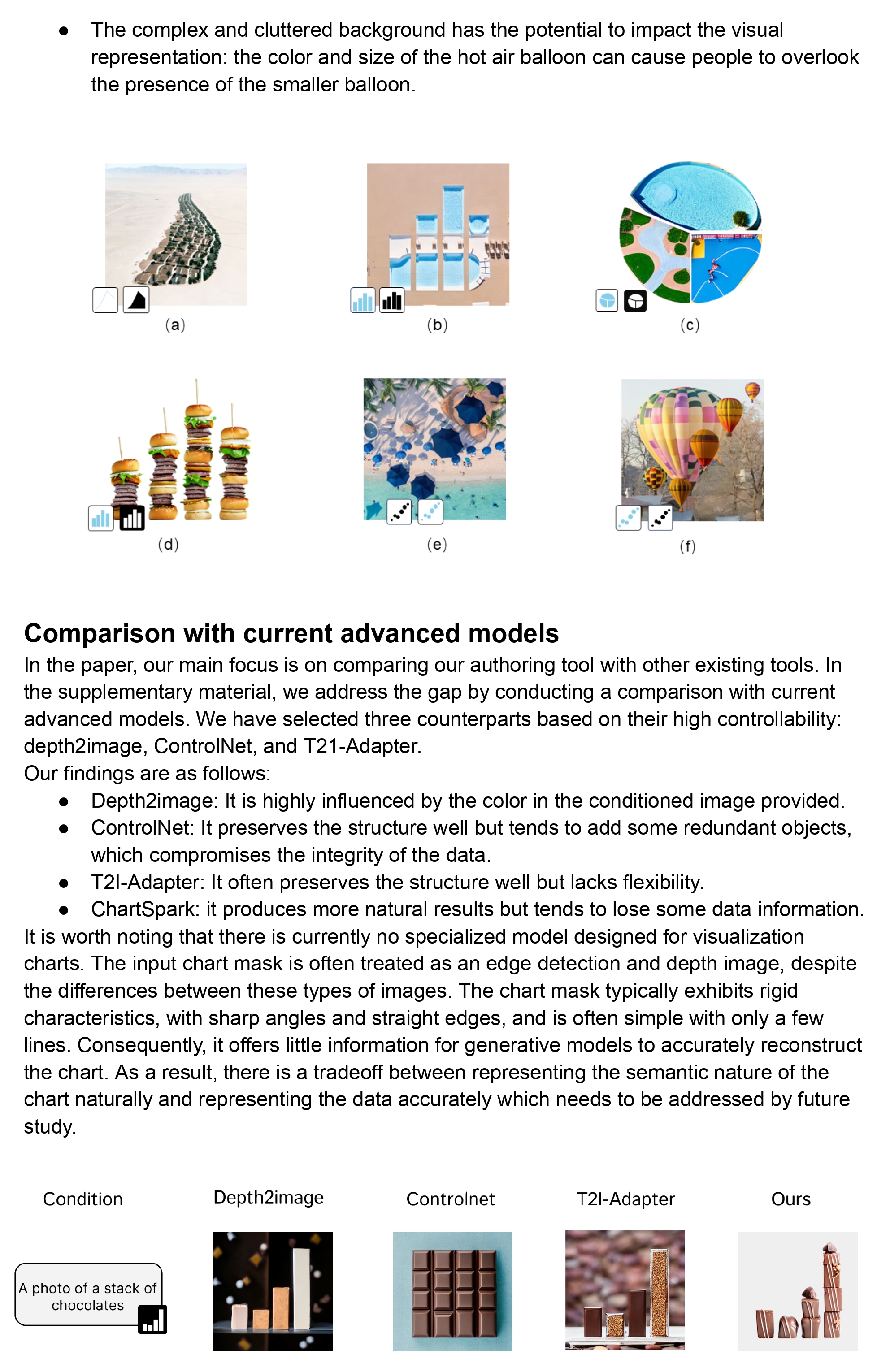}
    \vspace{6em}
    \caption{}
\end{figure*}

\begin{figure*}[t]
    \centering
    \includegraphics[width=0.9\linewidth]{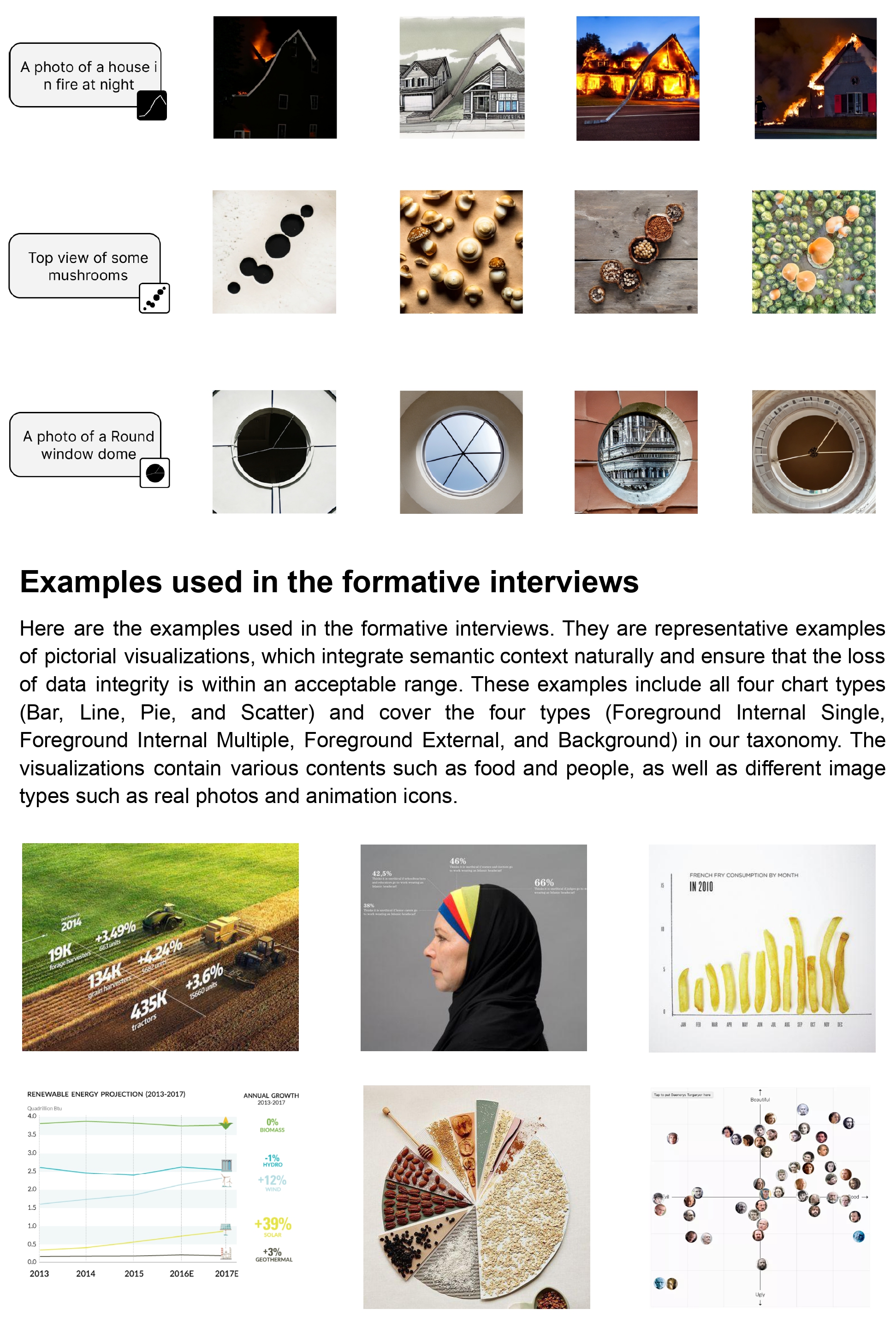}
    \vspace{6em}
    \caption{}
\end{figure*}

\end{document}

%% file: Latex/0_Abstract.tex
\abstract{
Pictorial visualization seamlessly integrates data and semantic context into visual representation, conveying complex information in an engaging and informative manner.
Extensive studies have been devoted to developing authoring tools to simplify the creation of pictorial visualizations.
However, mainstream works follow a retrieving-and-editing pipeline that heavily relies on retrieved visual elements from a dedicated corpus, which often compromise data integrity.
Text-guided generation methods are emerging, but may have limited applicability due to its predefined entities.
In this work, we propose \textit{ChartSpark}, a novel system that embeds semantic context into chart based on text-to-image generative model.
\textit{ChartSpark} generates pictorial visualizations conditioned on both semantic context conveyed in textual inputs and data information embedded in plain charts. The method is generic for both foreground and background pictorial generation, satisfying the design practices identified from an empirical research into existing pictorial visualizations.
We further develop an interactive visual interface that integrates a text analyzer, editing module, and evaluation module to enable users to generate, modify, and assess pictorial visualizations.
We experimentally demonstrate the usability of our tool, and conclude with a discussion of the potential of using text-to-image generative model combined with interactive interface for visualization design.

}

%% file: Latex/1_Introduction.tex
\section{Introduction}

% Greater aesthetic quality of a visual representation can indeed play a role in information intake [9], by increasing engagement [19, 29], comprehension [6] and memorability [10, 11].
% have a high capacity of piquing interest

% Visual embellishment and pictorial visualization are 

In the contemporary era of data explosion, visualization technology has become increasingly ubiquitous due to its ability to present complex data in a clear and vivid manner.
Pictorial visualization is one of essential techniques used to embed semantic context into chart, enhancing the visual representation hidden in the data\cite{holmes2022joyful,shi2022supporting,hartmann2017visualizing}.
The advantages of pictorial visualization are manifold, including improvements to long-term recall, user engagement and enjoyment, and information acquisition\cite{haroz2015isotype, borgo2012empirical,harrison2015infographic,moere2012evaluating}.
The aesthetic and practical values make the technique become widely adopted across a diverse range of applications, including advertisement, education, and entertainment\cite{hullman2011benefitting,moere2011role, borkin2015beyond, bateman2010useful,borkin2013makes}.

% Nevertheless, creating a visually appealing and informative chart requires a certain level of design expertise.
% Novice users may inadvertently include unnecessary and distracting elements in their charts, such as excessive gridlines, 3D effects, and decorative elements, commonly referred as ``\emph{chartjunk}'' \cite{tufte1985visual,few2011chartjunk}. 
% To address this issue, authoring tools that assist users in creating pictorial visualizations are emerging.
% Common tools can be generally divided into two categories.
% \emph{Rule-based} approaches, such as Piktochart\cite{pic} and DataShot\cite{wang_2020_datashot}, provide pre-designed templates or certain rules, along with interactive interfaces that allow for easy customization of pictorial visualizations.
% However, rule-based approaches have been criticized for being too rigid and limiting users' creativity.
% On the other hand, \emph{example-based} approaches, such as TimeLine\cite{chen2019towards}, apply machine learning techniques to extract design practices from existing visualizations and apply them to new designs.
% However, example-based approaches require a large number of well-designed examples to be effective.
Creating a visually appealing and informative chart requires a certain level of design expertise.
Recently, some studies approach the problem with more involved techniques and authoring tools that allow users to select graphic elements and then bind data to them. 
These approaches typically follow a retrieving-and-editing pipeline, which involves retrieving a suitable image from a large corpus and then adapting it to the visualization.
In the retrieving stage, the methods retrieve appropriate visual elements from a large image corpus, using the guidance that can be specified by a data file\cite{coelho2020infomages} or an image\cite{zhang2020dataquilt}.
The retrieved visual elements are then semi-automatically composed together, with user\red{s'} refinements enabled by an interactive interface, in the editing stage.
For instance, Infomages\cite{coelho2020infomages} retrieves an image containing the target visual \red{subject}, and then applies filling or overlaying techniques to adapt the data to fit the image.
% DataQuilt\cite{zhang2020dataquilt} provides an interface that enables users to extract visual elements from the retrieved image and composite them into a visualization chart.
Alternatively, the retrieved image can serve as a reference with style mimicked for new content\cite{qian2020retrieve,shi2022supporting}.
For instance, Vistylist\cite{shi2022supporting} decomposes the style of the reference chart into a tuple of color, font, \red{and} icon to guide the creation of new designs.
In the process, deep learning techniques (\eg,\cite{chen2019towards,qian2020retrieve}) can be employed, to facilitate the maintenance of the visual structure in the reference chart while extending it with new information.
\red{With the rise in popularity of language models, numerous works\cite{rashid2022text2chart,cui2019text,wang2022towards,lai2020automatic} have delved into the automatic generation of charts from text, bridging the gap between visualization and natural language.}
% For instance, generating visualizations related to proportions\cite{cui2019text} or highlighting important information\cite{wang2022towards}.
% Current studies utilize named entity recognition to enable natural language to perform various tasks, such as generating visualizations related to proportions\cite{cui2019text} or highlighting important information\cite{wang2022towards}.
% However, previous methods have several issues. 
% Firstly, retrieving an appropriate image that accurately reflects the target visualization object can be challenging.
% Secondly, editing the image to combine it with the chart is a complex task.
% Specifically, there is a compromise between fitting the chart into the image or adapting the chart to the image.
% Thirdly, these methods are constrained by a limited number of visual elements and a predefined set of text instructions
% From the user's perspective, all of these issues impede the creation process and entail design proficiency.

Nonetheless, these methods share some common issues.
First, retrieving an appropriate visual design that accurately aligns with the data requires a large-scale and high-quality image corpus, which is challenging to construct.
Studies in this direction often create new datasets, resulting in various datasets (e.g., \cite{cui2019text, lu_2020_vif, shi2022supporting, lan2021smile, morais2020showing}) that do not meet each other's requirements.
Second, editing  image to bind it with data is an error-prone task due to the trade-off between fitting and adapting the chart into the image.
Additionally, current text-guided methods are often limited by a predefined set of recognized entities and specific data types, such as proportion-related statistics \cite{cui2019text}.

To alleviate these limitations, we propose \textit{ChartSpark}, a novel pictorial visualization authoring tool that \red{embeds} semantic context into charts based on text-to-image generative model, while preserving important visual attributes such as trend.
We first extract the target visualized object \red{from the title of chart} through a text analyzer.
The generation process will then be guided by the object and its corresponding description provided by the user.
As an initial step towards utilizing generation techniques for creating visualization charts, we conducted a preliminary study that examines the different display formats of pictorial visualization. 
Drawing from this study, we categorized the embedded objects into foreground and background, with two types of generation methods for each, namely conditional and unconditional.
Unlike their definition in the natural image domain, we determine whether the method is conditional based on whether it takes into account chart information.
In addition, we offer an editing module for modifying and adding information, as well as an \red{evaluation} module for providing users with feedback on distortion perspectives.
By utilizing generation techniques instead of searching for images, we have three advantages over previous methods:
1) Eliminate the tediousness of searching and the possibility of not finding suitable images,
2) Allows for a more natural combination of semantic information and charts with robust generative ability,
3) Cover more visual elements and support flexible text instructions.
Therefore, our approach enables a text-guided generation and reduces the complexity of creating a pictorial visualization. In all, our \red{contributions} are three-fold:
% \begin{itemize}[leftmargin=*]
\begin{itemize}[leftmargin=*,topsep=0.25em]
    \setlength\itemsep{-0.45em}
  \item We create pictorial visualization based on generative model that can be conditioned by both semantic context and structure information.
  \item We construct an interface with text analyzer, editing module, and
evaluation module to assist user in creation process.
  \item The case study and expert interview demonstrate the effectiveness of our method.
\end{itemize}

%% file: Latex/2_RelatedWorks.tex
\section{Related Works}

\textbf{Pictorial Visualization}.
% Elicit More Empathy
% In align with
% 背景介绍
Pictorial visualization incorporates dedicated visual \red{semantics} to deliver complex data in an attention-grabbing manner.
\red{Except for the original form as ISOTYPE proposed by Otto and Marie Neurath\cite{hartmann2017visualizing},} there are various types of expressive visual forms emerging, including icon-based graph\cite{shi2022supporting}, proportion-related chart\cite{qian2020retrieve}, and timeline diagram\cite{chen2019towards}.
Compared to plain charts, the aesthetic appeal of pictorial visualization can \red{envision the topic\cite{burns2021designing}}, increase engagement\cite{harrison2015infographic,hullman2011benefitting,moere2011role}, and enhance information absorption\cite{borgo2012empirical,moere2012evaluating} and memorability\cite{haroz2015isotype, borkin2015beyond, bateman2010useful}.
Bateman \etal \cite{bateman2010useful} revealed that visually figurative charts could increase long-term recall by quantifying memorability attributes, including data-ink ratios and visual densities, without sacrificing interpretation accuracy.
Borkin et al.'s\cite{borkin2015beyond} ``beyond memorability'' study concluded that properly used pictograms do not interfere with understanding and can improve recognition.

% 负面:
% However, some visualization minimalists\cite{tufte1985visual,few2011chartjunk} argue that visual representations should be devoted to displaying data information maximally.
% % following the data-ink principle, that states that the amount of ink devoted to displaying data information should be maximized.
% Visually figurative charts can be distracting to users and reduce chart readability.
% These claims were supported by studies\cite{cleveland1984graphical, kosslyn1989understanding} that estimate the amount of quantitative information extracted from graphical perception.
% Despite the ongoing debate on ``chartjunk'', studies (\eg,\cite{few2011chartjunk, borkin2015beyond}) generally recognize the inherent benefits of visual embellishments, as long as they do not cause distractions or misinterpretations.
\red{These pictographs are hailed as ``\textit{charthelp}'' \cite{holmes2022joyful}, which arouses researchers' interest in exploring the design space for pictorial visualization.}
Some studies\cite{boy2017showing, morais2020showing} focus on expanding the design space and propose various design dimensions as guidelines for practice.
Recently, more efforts have been devoted to \red{investigating the embedding types of combining semantic context and data.}
\red{Byrne et al.\cite{byrne2019figurative} showed the role of figurative background in hybrid visualization as a detailed description of data content.
Some researchers\cite{ying2022metaglyph,zhang2020dataquilt} focus on replacing the data representation in the foreground of chart.
For instance, Ying et al.\cite{ying2022metaglyph} automated the glyph design that encodes visual metaphor.}
% 种类
% 吸取这些评价
% Thus based on findings in these studies, researchers in the field have been inspired to create tools that aid in the design of appropriately embellished charts. We discuss these studies and tools below.
% 我们的关注到了什么题
% 关注到了他的重要性
In line with these research directions, \red{our work aims to categorize and organize the various embedding types from a large pictorial visualization corpus and endeavor to support flexible generation.}

\vspace{1mm}
\noindent
\textbf{Visualization Authoring Tool}.
% 总说：针对pictorial visualiztion, 很多tool
Numerous authoring tools have been devised to streamline the creation of pictorial visualizations.
% 用户去创造
Previous works\cite{kim2016data,xia2018dataink, wang2018infonice} focus on mapping visual elements to data using suitable visual encoding channels, which require the expertise of a trained designer.
% dataguide，dataink提供了一个好的工具去bind用户设计的图和数据结合
% [Dataink] provides tools for direct manipulations that enable users to bind their graphics with data, which leads to flexible and creative visualization. 
% [Dataguide] offers guides to encode data into expressive graphics while  maintaining data integrity. 
% to alleviate 需要画需要设计，很多工作都follow “seeking-and-editing” 的模式，先xxx再xxx。
% Both tools give users a high degree of flexibility in creating visualizations, but requires a significant level of proficiency in design principles and techniques.
To simplify the process of designing visual elements, many approaches implement a retrieving-and-editing pipeline to draw inspiration from a vast repository of dedicated resources.
% The process involves finding an image that satisfies certain properties and then adapting it into a chart using different editing methods.
% 按照seek的对象我们可以分成以下两类
There are two types of designs based on retrieval: content-based design and example-based design.
% content-based design --- 把图中语义信息相同的物体抠出来reuse
Content-based designs\cite{zhang2020dataquilt,coelho2020infomages} involve searching for an image that shares semantic context with the original chart, and merging its visual representation with the chart. 
%% [infomages] [dataquilt]
For instance, Infomages\cite{coelho2020infomages} retrieves an image that contains the target visualization object and subsequently applies filling or overlay techniques to adapt the chart to fit the image. 
These methods can be time-consuming and compromise data integrity due to manual collation.
% [DataQuilt] has developed an interface that enables users to extract visual elements from the retrieved image and composite them into a visualization chart.
% Example-based design --- 用style
Example-based designs\cite{shi2022supporting,qian2020retrieve,chen2019towards,wang_2020_datashot} focus on retrieving a well-crafted chart as a reference and emulating its style for creating new content.
%% [Vistylist] [timeline] [retreive-adapt]
Vistylist\cite{shi2022supporting} disentangles the style and elements from a retrieved example and applies them to a new chart.
% Deep learning techniques are used in other studies to extract structures from reference charts and apply them to new content, specifically for certain visualization types such as proportion-type charts in [retrieve-adapt] and timeline charts in [timeline].
% color schemes, visual expression
% 问题：however， seek 和 edit 都很难
% However, it can be challenging to either find the appropriate image or adapt it to the intended chart using the seeking-and-editing pipeline.
% 现今一些工作开始自动化流程，用text引导chart的生成。

Despite their apparent simplicity, these approaches still require visualization design expertise that may be challenging for novice users.
Some recent works \red{decrease the barrier of creating pictographs} by employing language models to guide the creation process\cite{cui2019text} or highlight the intended information\cite{wang2022towards}.
% [text-to-vis] 针对proportion
% For example, [text-to-vis] generates visualizations according to proportion-related statement while
% [vistalk] 对chart进行modify， including highlight， addxx
% [vistalk] can modify charts by the given instructions including highlighting important data and adding visual elements.
% 问题：predefined set of text instructions
Nonetheless, such methods heavily depend on a predetermined set of recognition entities and specific data types, such as only being able to recognize proportion-related statistics\cite{cui2019text}.
% 我们：
% 用生成的方式解决了seeking难+editing难的问题，并且不通过预设text去限制用户。
% In this work, we further investigate using text to enhance pictorial visualization design, supporting flexible textual inputs.
% Meanwhile, we 
In this work, \red{we develop an authoring tool that simplifies the creation process, mitigating challenges associated with retrieving appropriate images and potential errors when manually editing the visual design.
Rather than relying on predefined recognition entities, our tool supports flexible text input to accommodate the diverse requirements.}

% In this work, we take a different approach by utilizing a \emph{text-to-image} generative model, instead of language models relying on structured statements, to enhance pictorial visualization design.
% With the generative model, users have the flexibility to provide textual inputs, and can avoid the challenges associated with finding appropriate images and the potential for error when manually editing the visual design.
% In this study, we utilize the text-guided generation technique to xx, streamlining the retrieval-and-editing process and improving the flexibility of text input.

% In our work, we employ a generation technique to a seeking-and-editing pipeline, which enhances the embedding of semantic context into charts by avoiding the challenges of finding suitable images and the potential loss of data integrity during the editing process.
% Moreover, we offer users more flexible text input options, without relying on a predefined set of text patterns.
% 加油加油！穗子最棒！
% 哈哈哈，好滴，谢谢妹宝~

\vspace{1mm}
\noindent
\textbf{Text-guided Image Generation}.
Text-to-image diffusion models have demonstrated great success in producing high-quality images, surpassing previous mainstream GAN models\cite{karras2019style,gal2022stylegan} and autoregressive models~\cite{yu2022scaling,ramesh2021zero}.
% DDPM DDIM 等出现改变了生成模型，why outperform
Pilot works\cite{ho2020denoising,dhariwal2021diffusion,song2020denoising} use a hierarchy of denoising autoencoders with desirable properties such as a stationary training objective and good generalization, laying a solid foundation for following research.
% 文字生成
To enable text conditions, Rombach et al.\cite{rombach2022high} incorporate cross-attention layers\cite{hertz2022prompt,chefer2023attend}.
% and operate in latent space, solving the issue of costly inference.
Stable Diffusion\cite{ramesh2022hierarchical} achieves zero-shot performance through the utilization of CLIP's multi-modality space\cite{radford2021learning}, while Imagen\cite{saharia2022photorealistic} discovers that pretrained language models can serve as a practical text encoder for generation purposes.
However, plain text-guided generation lacks control, leading to the development of models that emphasize controllable manipulation.
% near-optimal point between complexity reduction and detail preservation, greatly boosting visual fidelity.
% 可控生成 controllable manipulation
Some studies\cite{gal2022image, ruiz2022dreambooth} finetune generative models with a set of images to maintain consistent style in newly synthesized images.
% Other studies\cite{hertz2022prompt,chefer2023attend} explore the cross-attention mechanism and inject semantic information throughout the diffusion process.
Recent advancements in conditional models\cite{li2023gligen,zhang2023adding}, which incorporate text as well as additional grounding concepts such as bounding boxes and keypoints, have made it feasible to tackle the issues of layout and composition.

% 我们的方法，也是用attention去达到可控生成，与上述方法不同，我们还融合了图表的信息
The methods above focus on natural images, while we are pioneering the use of text-to-image generative models in visualization.
\red{Some works\cite{xiao_2023_wytiwyr, wu2023viz2viz,Schetinger2023DoomDeliciousnessChallenges} also discuss the promise of applying vision-language models in visualization. Wu et al.\cite{wu2023viz2viz} devised several pipelines but are limited by presetting an image with arranged semantic objects and color, and hard to evaluate the data integrity. 
We propose a universal generative pipeline that includes comprehensive embedding types and generation methods for several different types of charts. This pipeline offers an end-to-end generation process, which frees users from manually designing an initial chart. Furthermore, we integrate the modification and evaluation of the generated charts to enhance performance.}
% This is nevertheless a challenging task, as the generated pictorial visualizations must maintain data integrity while also accurately reflecting the semantic context.
% To address this challenge, we integrate chart information into the textual guidance through the use of attention mechanisms.
% In contrast to the aforementioned techniques used for natural images, we are taking the first step towards utilizing generative models in the field of visualization. Moreover, we are including chart data to strengthen the accuracy of semantic context and data integrity through the implementation of attention mechanisms.

%% file: Latex/3_Preliminary.tex
\section{Preliminary Study}
We conducted a preliminary study to comprehend crafting a pictorial representation.
% We acquire knowledge about the general workflow, the concern, and expectation about using generative models, and design requirements through formative interviews, and collect and summarize typical embedding types regarding integrating visual elements and charts.
From the formative interviews, we learned the design workflow (Sect.~\ref{ssec:interview}), and received concerns and expectations that are summarized as design requirements. 
We collected a corpus of pictorial visualizations and examined the design patterns (Sect.~\ref{ssec:analysis}). 
% summarize various embedding types for integrating visual elements and charts that are typically used. and 

% 得到用户创造一个pictorial vis 的流程，requirements
\subsection{Formative Interviews}\label{ssec:interview}
% 目的
To ensure our authoring tool is accessible to a wide range of users, we conducted formative interviews with people from three different backgrounds: two artists (A1, A2) \raisebox{-.2\height}{\includegraphics[width=0.32cm]{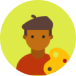}} who mostly use Adobe Illustrator and P5.js in daily work \red{and have infographic design experience for art track of some conferences}, two visualization experts (V1, V2) \raisebox{-.2\height}{\includegraphics[width=0.32cm]{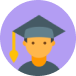}} who have around \red{3-year} data analysis and dashboard design experience, and two individuals (P1, P2) \raisebox{-.2\height}{\includegraphics[width=0.32cm]{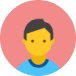}} without relevant training in art or visualization.
\red{Both artists and experts are familiar with pictorial visualization, as they have learned, seen, and created this type of visual representation. The other two individuals were unfamiliar with the professional term "pictorial visualization," but remembered seeing this type of visualization in daily life after we showed some examples.}

% %% pipeline
% From the interviews, a general workflow in creating pictorial visualization was summarised.
% %% 不同的人的看法 a,b,c
% Also, opinions from participants about pictorial visualization were varied across different disciplines.
% 采访时间
Each participant was interviewed individually, and the length of each interview varied from 30 minutes to an hour.
% 采访的流程：1. 展示几个typical 的pic，2.给他们1组数据about 全球沙漠面积变化状况，要求他们讲出将数据变成pic，他们的操作流程。3.针对这个过程，询问他们三个问题：1) 2) 3) 用生成模型去做，有什么担忧和期待？concerns and expectation.
% 要求think aloud，画草图。
The interview consisted of three stages.
First, we introduced the concept of pictorial visualization and presented several examples. \red{Please see the Supplementary for detail.}
During the second stage, participants were provided with data on the global change in a desert area, which included an x-axis for time, a y-axis for the area, and a title. They were then instructed to describe their design workflow using rough sketches \red{on iPad} while vocalizing their thought process.
\red{It's flexible and allows for easy import of visual elements into the sketch, as well as manipulation such as rotation and scaling.}
\red{Throughout the process, we follow the think-aloud protocol, recording each step the user takes in the sketch.}
Finally, we asked the participants the following questions: 
1) What are the key steps involved in creating a pictorial visualization?
2) Which step in the pictorial visualization creation process do you find the most challenging, and why? 3) What expectations and concerns would you have if a generative model were involved in creating pictorial visualizations?

\subsubsection{General Workflow}
\begin{figure}
    \centering
    \includegraphics[width=\linewidth]{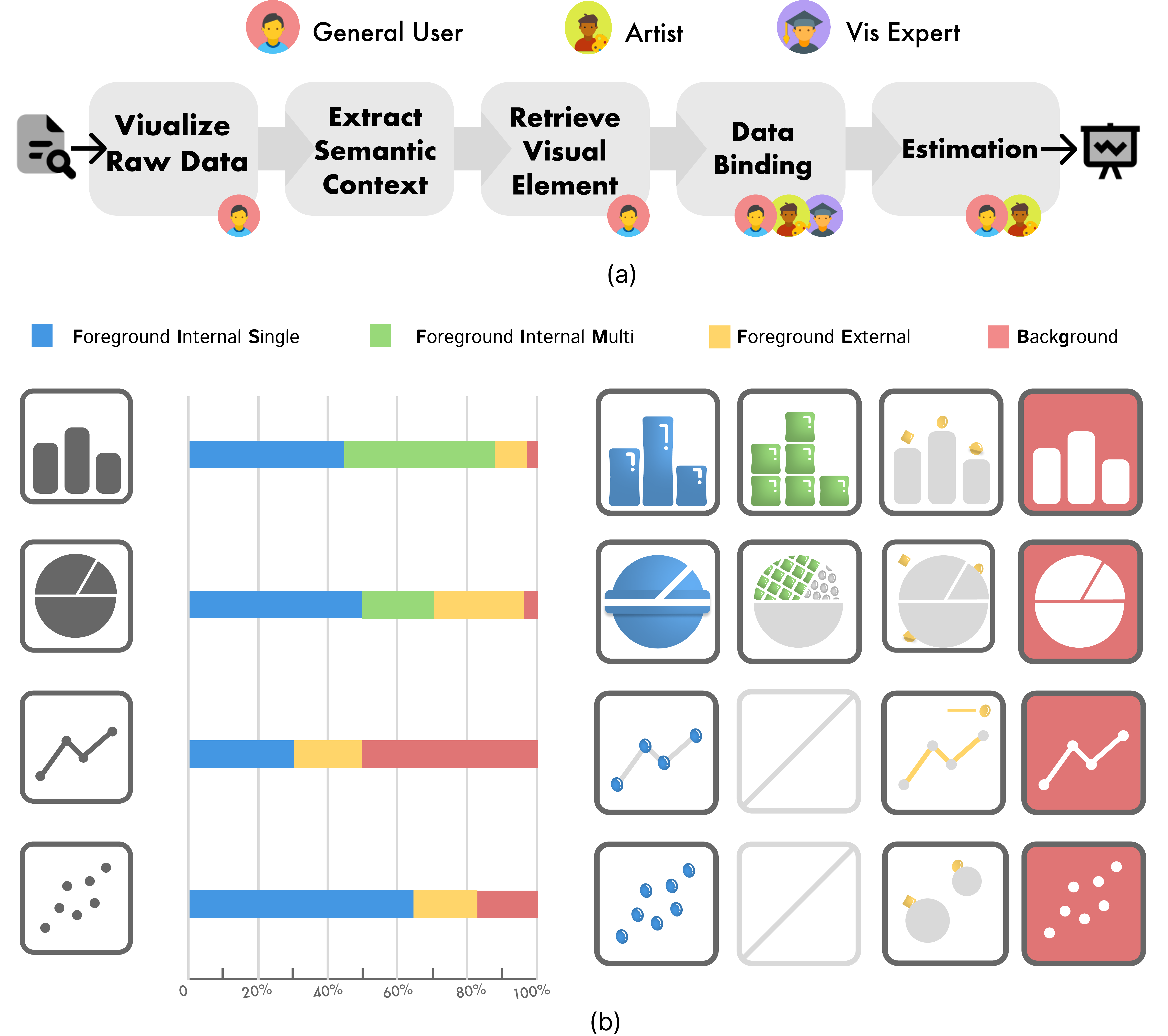}
    \vspace{-1.5em}
    \caption{\textbf{Analysis of preliminary study.} (a) General workflow of creating pictorial visualizations contains five stages. Participants with different backgrounds have encountered difficulties on the stage with indications. (b) Taxonomy of common design patterns for different chat types. We represent the percentage of each design pattern for each chart type in the corpus. For illustrating the representation of each design patterns, we use several types of candy to represent.}
    \label{fig:flowchart1}
    \vspace{-1.6em}
\end{figure}

% observation 就是发现大家都遵从这么一个流程，如图所示。
The participants' workflows in creating pictorial visualizations revealed that there was a general pattern to the process, as shown in Fig.~\ref{fig:flowchart1} (a). 
% 简要描述流程
Firstly, participants recognized the theme and data features by reading the title and observing the trend to quickly comprehend the data.
Next, they draw or retrieve some visual elements reflecting the theme. 
Once the visual elements were confirmed, participants attempted to bind the data with these elements using techniques such as rotation, scaling, deformation, and stretching.
At the end of the creation process, some participants evaluated the final design to ensure the accuracy of data binding and avoid the error of manual manipulation.

% 共性问题：data binding 的过程非常繁琐和富有挑战性
Based on the interviews, we found that participants all mentioned that it was tedious and challenging to bind data with visual elements.
%% 艺术家，vis expert： 繁琐，因为对每一个element都要做相似的操作， 其中艺术家提到“更换素材意味着重头开始，不利于创作迭代”
\textit{``I need to repeat the same operation for each single element,''} A2 commented, \textit{``If I want to change visual materials, I have to start my work from the scratch. It is not good for creation iteration.''}
%% commonpublic： “抠图，去除锯齿等需要专业软件，且针对chart的形状做deformation是很复杂的”
Meanwhile, P1 noted that \textit{``The image cutout and anti-aliasing need to use professional software that is hard to use. It is also complicated to deform the visual element to adjust the shape of charts.''}
% 在采访过程中，也发现大家有不同的侧重点
% For participants with diverse backgrounds, 她们有不同的侧重点。
% 差异性：
%% common public： 认为找主题和找图片很麻烦。“特别是一些抽象的词汇，很难找到一个具象的element去表达”
Furthermore, both P1 and P2 stated that it was difficult to find images that matched with the theme. \textit{``I can barely find any concrete elements, especially for abstract vocabulary,''} mentioned by P2. However, neither artists nor visualization experts thought that would be a potential issue.
%% estimation： 分vis expert和common public来说。
As for evaluating visualization performance, visualization experts raised up the concern about the visual distortion and data integrity compared with the original chart.

\subsubsection{Design Requirements}
% concerns and expectation.
% 总说: 在讨论完convential 流程之后，大家对将生成模型纳入这个过程产生了强烈的兴趣，并提出了自己的担忧和期望。
Besides the discussion about the conventional workflow, participants also expressed their expectations and concerns about involving generative models in creating pictorial visualizations.
% R1在开始创作之前，他们认为捕捉到合适的语义信息和显著的数据特征非常重要，尤其是在利用AI生成的语境下。 
All participants believed that it was essential to capture appropriate semantic context and significant data features before starting to design.
%% 他们都是先画个反映数据的chart sketch，再这个基础上添加语义信息。
% They sketched a chart that reflected data at the beginning of their design process, which provided a solid base for adding the visual elements with relevant semantic context.
%% 当告知他们AI可以辅助他们生成pict，他们认为这种preview raw data更加有必要去判断AI生成是否faithful。
They emphasized the necessity of \red{offering visualization preview for raw data} so that they could evaluate if the portrayal of data by the generative model was faithful. 
% R2 R3
% 希望control， customization 
Participants hoped that the generative model could allow flexible customization in design. 
Specifically, some participants wanted a controllable generative result including its shape and color while others wanted a large variety of styles in visual elements. 
% 看把some people替换成哪些人吧
% 生成图片的同时integrate visual attributes from chart. V2 "现在的文生图模型效果很好，但针对pictorial visualization，还要考虑图表的trend，bar 高度等visual encoding信息"
Moreover, participants expected generated image can integrate both visual elements and data simultaneously. \textit{``Recent text-to-picture models work well in general,''} V2 acknowledged, \textit{``However, particularly for pictorial visualization, we must consider information encoded by visual channel like the trend for line plot and height for bar chart.''}
% R4 
% 在作品快完成前，希望有一个estimation机制能够帮助用户评估最后生成的质量，从distorion和disharmony角度。
All participants anticipated an evaluation module to validate the quality of their works once the visualization was completed, which provided the visual distortion from the original data including height, area, and angle.

% There are DR based on participants' concers (R1 R4) and expectations (R2 R3), 按设计流程总结如下:
Following the workflow, there are four design requirements based on participants' concerns (R1 and R4) and expectations (R2 and R3).

% \begin{enumerate}[leftmargin=*, label=\textbf{R\arabic*.}]
\begin{enumerate}[leftmargin=*,topsep=0.25em, label=\textbf{R\arabic*.}]
    \setlength\itemsep{-0.45em}
\item \textbf{Preview data and theme.} Visualizing the raw data and obtaining semantic description before starting visualization design.

\item \textbf{Personalize visual element.} Customizing pictorial visualization, such as color and shape by controllable manipulation, while expanding design space with various styles of visual elements. 

% \item \textbf{Simplify data binding.} 
\item \textbf{Embed semantic context into chart.} Integrating the visual element and data automatically and naturally, while supporting flexible embedding methods for the semantic context.
%let visual element automatically and naturally fit the data, such as trend, height, etc.

\item \textbf{Evaluate the performance.} Evaluating visualization design in visual distortion, which indicates the loss of data integrity.
\end{enumerate}

% 对现有的进行分析 taxonomy
\subsection{Pictorial Visualization Corpus}
\label{ssec:analysis}
%%%%%%%%%%%%%%% 冲啊！！！！！！！！！！！！！！！ 
Our preliminary study also unveils systematic patterns for integrating semantic context with data in pictorial visualizations. To collect the sample, we drew upon datasets provided by prior research \cite{coelho2020infomages,shi2022supporting, lan2021smile} and manually selected some typical charts, resulting in 587 charts to constitute part of our data.
As some of the collected pictographs overly focused on a specific type such as icon-based\cite{shi2022supporting}, we also retrieved additional examples from Pinterest and Google to supplement our corpus, comprising 869 samples.
We then classified them based on embedding types.
To minimize individual judgment bias, each visualization was appraised and categorized by two authors, with a double-check process.
% We analyzed the collected data and summarized the common design patterns into a taxonomy.
As depicted in Fig.~\ref{fig:flowchart1} (b), we employ various shapes of \red{candy icons} to represent this taxonomy.

\begin{itemize}[leftmargin=*,topsep=0.25em]
    \setlength\itemsep{-0.45em}
\item \textbf{Foreground Internal Single} (395, 45.5\%). 
The semantic context is embedded in foreground, encoding the visual element in a single manner as the chart itself. 
Examples include rectangular candies for bars in a bar chart and round candies for points in a scatter plot.

\item \textbf{Foreground Internal Multiple} (290, 33.4\%). 
The semantic context is embedded in foreground, encoding the visual element in a multiple manner as the chart itself. 
For example, a stack of the same candies to form a bar in a bar chart. 

\item \textbf{Foreground External} (121, 13.9\%). 
The semantic context is embedded in foreground, encoding the visual element in a single manner and locating it externally. 
For instance, there is a candy next to each sector in a pie chart.

\item \textbf{Background} (63, 7.2\%).
The semantic context is embedded by visual elements in the background. 
\red{Some examples use an image as a background to depict the theme.}
\end{itemize}
%% 怎么做的
%% 占比  
%% chart 差异
Overall, the design pattern concerning the embedded object can be divided into foreground and background.
While the visual representations are diverse, we observe a significant difference in these examples: whether the data can be reflected in the visual element. \red{For foreground external representations, the data information is often not included.
In contrast, some examples in the other three types would comply with the data's inherent trend or magnitude.}

% If so, the semantic information and data information will be encoded together within the element, and the element would comply with the data's inherent trend or magnitude.
% While embedding semantic context into visual elements, some embed objects also capture the chart information for certain embedding type. 
% For example, Foreground Internal Single preserves chart information, such as height in bar chart and trend in line chart when generating visual elements, Foreground External does not consider chart information. 

%% file: Latex/4_Method.tex
\section{Method}
\subsection{Overview}
\begin{figure}[t]
    \centering
    \includegraphics[width=\linewidth]{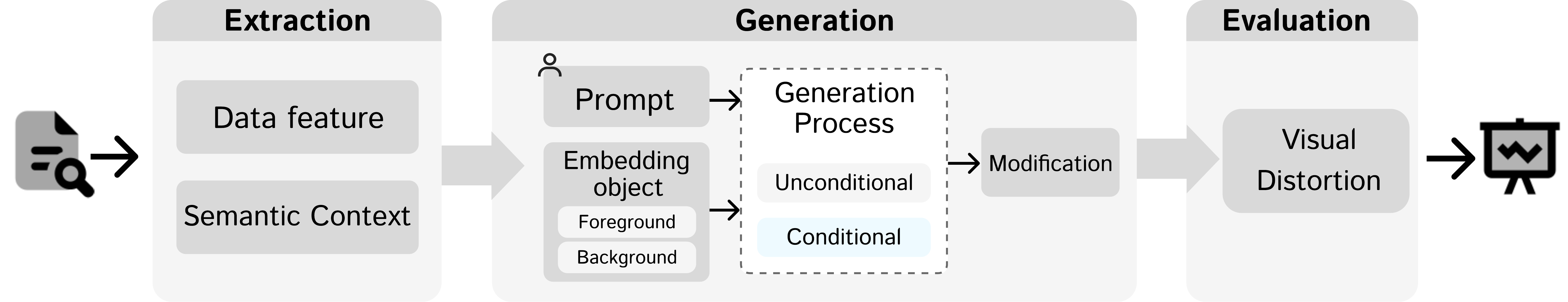}
    \caption{\textbf{The 3-stage framework of \textit{ChartSpark}.} The extraction stage provides users with data features and semantic context. The generation stage produces visual elements by the input prompt and selected method with a final refinement. The evaluation stage evaluates the generated visualization based on distortion.}
    \label{fig:flowchart2}
     \vspace{-1.4em}
\end{figure}
\begin{figure*}[t]
    \centering
    \includegraphics[width=\linewidth]{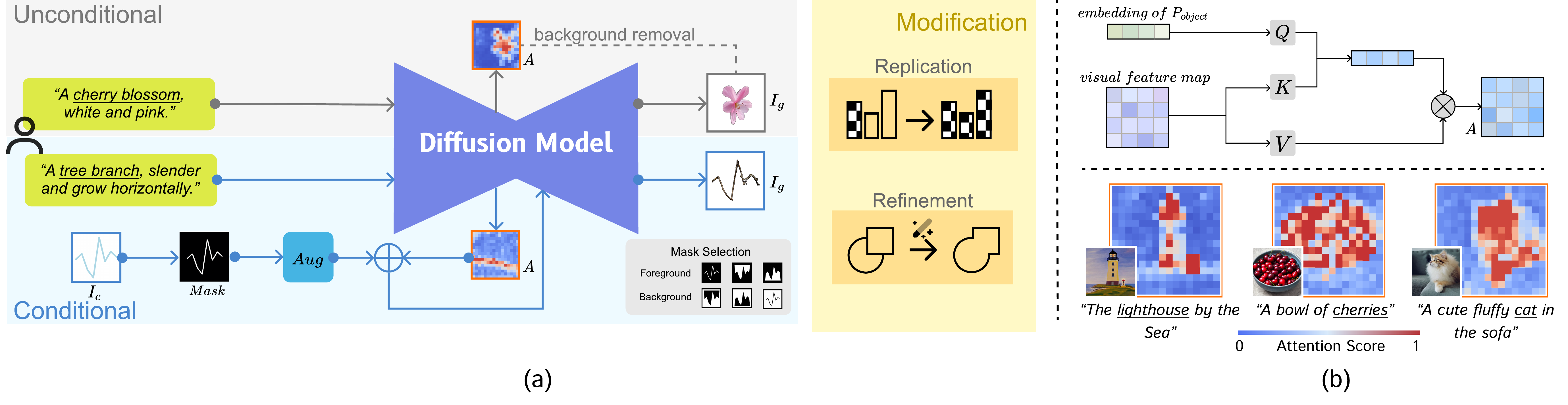}
     \vspace{-1.8em}
    \caption{\textbf{The process for unconditional and conditional generation for foreground and background.} (a) Given a prompt, the generative model generates relevant visual elements by unconditional or conditional method, then the generated visualization can be edited through replication and refinement. (b) Internal mechanism of incorporating an image and textual input into the attention map.}
    \label{fig:method}
    \vspace{-1.5em}
\end{figure*}
The proposed \textit{ChartSpark} framework's workflow is depicted in Fig.~\ref{fig:flowchart2}, comprising three primary stages.
% 介绍一下我们的流程, 对应一下R
In the initial stage, data features are visualized and semantic context is extracted from the raw data, offering users a visual preview and thematic topic to enhance their comprehension of the data (\textbf{R1}).
Subsequently, users employ prompt-driven generation to acquire visual elements, embedding the semantic context into the foreground or background in either a conditional or unconditional manner (\textbf{R2, R3}).
Ultimately, the evaluation module furnishes a suggestion mechanism to indicate data distortion (\textbf{R4}).
% 对比一下以前的
In comparison to the workflow depicted in Fig.~\ref{fig:flowchart1} (a), \textit{ChartSpark} streamlines the process by supplanting the retrieval and data binding stages with the generation, which mitigates the inconvenience of retrieving visual elements and integrating them with charts.
Instead, users provide prompts regarding their preferred style and embedding technique to steer this automatic generation.
% 三明治结构
The \textit{ChartSpark} framework features a sandwich structure, with the first and last stages ensuring optimization of the middle stage's performance and augmenting the faithfulness in data and expressiveness in visual representation.
The preview presented in the initial stage enables users to intuitively discern the potential distortion of the generated chart, while the evaluation in the final stage delivers more accurate values and explicitly visualizes the error.

\subsection{Feature Extraction}
% visual信息 和 semantic信息
% text and numeriacal data
% Since pictorial visualization fuses both numerical data and textual information, we extract the data feature and semantic context from raw data, respectively.
User input is tabular data in JSON or CSV format, typically including numerical values for describing x- and y-axis, and a title optionally for conveying the theme.
To provide users with a preview of their data, we extract both data features and semantic context, respectively.

\noindent \textbf{Data feature.}
% To make user quickly perceive the overall appearance of raw data, We visualize it and offer multiple chart types for them to choose.
% Moreover, such data preview can serve as a hinter to hint the 偏差 in the following generation process.
We offer several chart types from which users can choose to visualize (Fig.~\ref{fig:interface} ($A_{1}$)).
To help them efficiently identify patterns and trends concealed within the raw data, \red{we initially visualize them as plain charts} (Fig.~\ref{fig:interface} ($A_{2}$)).
These data previews function as an indicator to detect any deviations in the subsequent generation process.
Furthermore, we extract the data annotation encompassing the $x$ and $y$ axes and the title in an SVG format, making it editable on the interface.

\noindent \textbf{Semantic context.}
To extract semantic context, we employ a two-step approach, namely keyword extraction and relevant word retrieval.
Initially, we use MPNet\cite{song2020mpnet}, a pretrained model with a sentence transformer, to extract keywords from the title provided in the raw data.
Moreover, we provide relevant words to stimulate users' creativity, especially those with limited design expertise, according to Fig.~\ref{fig:flowchart1} (a). 
We estimate word similarity using Word2Vec\cite{mikolov2013efficient} and convert words to vectors.
The corpus for retrieval is the English Wikipedia Dump from November 2021, which consists of 199,430 words, each with a 300-dimensional vector representation.
\red{However, the retrieved results based solely on similarity may contain proper nouns, such as "NASDAQ", a brand of the financial institute.
Using such proper nouns in charts may not be easily recognizable.
Instead, we expect general concepts to make pictorial visualization more memorable.
Therefore, the frequency of occurrence of the word in the corpus is leveraged to rank the order of the retrieved results, see Fig.~\ref{fig:interface} ($A_{3}$).}

% 1. similar words
% 通过距离找相似词，word2vec
% 199430 words
% 300 dimension
% Following the \cite{kutuzov2017word}
% 2. computer the frequency of these words

\subsection{Generation}
% 符号定义
% Let I be an image that was generated by a text-guided diffusion model using the text prompt P and a random seed s. Our goal is to edit I, using only the guidance of an edited prompt P∗, in order to get an edited image I∗ that maintains the content and structure of the original image but corresponds to the edited prompt. For example, consider an image generated from the prompt “my new bicycle”, and assume that the user wants to edit the color of the bicycle or replace it with a scooter while preserving the appearance and structure of the original image. An intuitive interface for the user is to directly change the text prompt by further describing the appearance of the bike, or replacing it with another word, respectively. As opposed to previous works, we wish to avoid relying on any user-defined mask to assist or signify where the edit should occur. A simple, but unsuccessful attempt is to fix the internal randomness and regenerate using the edited text prompt.

Based on the preliminary study, we identify four embedding types, along with foreground and background as the two main embedding objects.
A key observation is that some examples only require visual embellishment containing semantic context, while others need to comply with the inherent data to make it part of the chart itself.
In light of this, we devise a generation methodology that employs unconditional and conditional approaches.
The fundamental distinction between these approaches hinges on whether the chart information is factored into the generation process.
As illustrated in Fig.~\ref{fig:method}(a), the generation stage consists of three core modules.
The \emph{unconditional module} adopts the fundamental text-to-image diffusion model, subsequently generating a corresponding visual element.
For the \emph{conditional module}, we inject the chart image into the attention map, serving as guidance for the ensuing generation process, as shown in Fig.~\ref{fig:method} (b).
Lastly, the \emph{modification module} is tasked with \red{replication} and refinement to accommodate the four embedding types and enhance its details.

% Our key observation is that the structure and appearance of the generated image depend not only on the random seed, but also on the interaction between the pixels to the text embedding through the diffusion process. By modifying the pixel-to-text interaction that occurs in cross-attention layers, we provide Prompt-to-Prompt image editing capabilities. More specifically, injecting the crossattention maps of the input image I enables us to preserve the original composition and structure. In Section 3.1, we review how cross attention is used, and in Section 3.2, we describe how to exploit the cross-attention for editing. Self-attention is discussed in section 3.3. For background on diffusion models, refer to appendix B.

\subsubsection{Unconditional Generation}
% 介绍背景
% Diffusion-based generation methods are gaining popularity due to their outstanding performance.
Diffusion-based generation methods outperform previous generation methods in the quality of generated images and semantic comprehension. 
In this work, we develop our framework based on the Frozen Latent Diffusion Model (LDM) \cite{rombach2022high}.
Below, we outline the core structure and generation process of LDM to provide some preliminary knowledge.
Similar to the previous diffusion models\cite{ho2020denoising,dhariwal2021diffusion,song2020denoising}, LDM follows a forward process that incorporates Gaussian noise using a Markov process, and a reverse process that denoises to recover the original distribution from reconstructing the image.
However, LDM distinguishes itself from other diffusion models by employing compressed and low-dimensional latent space instead of pixel-based diffusion, thereby reducing computational costs for inference.
The LDM architecture includes an autoencoder that converts pixel images into latent representations and a UNet that predicts noise and performs denoising in the latent space.
To enable text-guided generation, LDM enhances the underlying UNet backbone with a cross-attention mechanism, facilitating the integration of multimodal guidance.

% SD is a two-stage diffusion model, which contains an autoencoder and an UNet denoiser. In the first stage, SD trained an autoencoder, which can convert natural images X0 into latent space and then reconstruct them. In the second stage, SD trained a modified UNet [20] denoiser to directly perform denoising in the latent space.
%% 两个阶段
% \begin{figure}
%     \centering
%     \includegraphics[width=\linewidth]{figs/unconditional.pdf}
%     \caption{\textbf{Cross-attention mechanism.} Top: visual feature and}
%     \label{fig:uncon}
%     \vspace{-2em}
% \end{figure}

% foreground 需要把东西给弄出来
\noindent \textbf{Foreground.}
% 背景：
In our preliminary analysis of existing pictorial visualization, the foreground is the common object to embed the semantic context and can exhibit various representations.
The unconditional generation can produce visual elements to embellish the chart, which closely matches the semantics but does not contain information about underlying data.
%与语义信息密切相关但是不包含数据信息。
% 总说 what， goal
We achieve this by utilizing the prompt-driven method.
The text prompt $P$ provided by the user consists of an object $P_{obj}$ and its corresponding description $P_{des}$.
In Fig.~\ref{fig:method} (a), the term with an underline represents $P_{obj}$, while the term without an underline represents $P_{des}$.
Given the generated image $I_{g}$, our objective is to extract the visual element related to the semantic context $P_{obj}$ from the image.
To accomplish this, we use cross-attention between object $P_{obj}$ and $I_{g}$ to locate the target region, and then remove the background to obtain $I_{obj}$. 
As shown in the top of Fig.~\ref{fig:method} (b), we obtain the visual feature map $V$ of the generated image from the autoencoder and embedding of $P_{obj}$.
Next, we use linear projections to transform them into $Q$ and $K$. We then multiply $Q$ and $K$ to obtain the attention score, which is subsequently multiplied with $V$ to generate the final attention map.
In summary, the process can be described as follows:
% \vspace{-0.5em}
\begin{equation}
\label{eq1}
\vspace{-1em}
A(Q,K,V) = Softmax(\frac{QK^{T}}{\sqrt{d}})\cdot V,
\end{equation}
% \vspace{-0.5em}
where the $d$ represents the dimension of the latent projection dimension of $Q$ and $K$, and the $Softmax$ function is utilized to normalize the attention score.
As shown in the bottom of Fig.~\ref{fig:method} (b), the attention score is directly proportional to the strength of the relevance between the image and text.
As a result, we can extract the object of interest from a cluttered background by comparing pixel differences.
To accomplish this, we first calculate the threshold to distinguish the object and background, obtaining a mask.
Next, we perform a pixel-wise comparison at the corresponding positions in $I_{g}$ to obtain a rough object region, denoted as $I_{obj}$. Lastly, to achieve a more refined result, we utilize ISNet\cite{qin2022highly}, a state-of-the-art segmentation neural network, to eliminate redundant information.
The process can be described through the following equations:
% \vspace{-0.5em}
\begin{equation}
\label{eq2}
M = \mathbf{I}[A_{ij} > \frac{\sum A_{ij}}{N^2}],
\end{equation}
\vspace{-1em}
\begin{equation}
\label{eq3}
I_{obj} = f_{upsample} (M) \odot I_{g},
\end{equation}
\vspace{-1em}
\begin{equation}
\label{eq4}
I_{obj}' = R(I_{obj}) ,
\end{equation}
% \vspace{-0.5em}
where $\mathbf{I}[.]$ is the element-wise indicator function on the matrix, $N^2$ represents the total number of pixels in attention map $A$, and $M$ is calculated as a matrix with a value of 1 at the object's location and 0 elsewhere. 
Since $M$ has the same dimensions as $A$, we employ an upsample technique to resize its shape to match that of $I_{g}$. $R$ represents the redundant information removal operation.
The symmetric and hierarchical structure of the UNet involves both downsampling and upsampling, resulting in cross-attention layers being present at different resolutions. In our experiments, we observed that the middle layer exhibited better performance, and empirically set the $N$ as 16.

\noindent \textbf{Background.}
% The preliminary study revealed that, in most cases, embedding semantic context into the background did not include data information according to user preferences. However, in the case of line charts with limited space in the foreground (such as the line itself), it is preferable to use the background to convey both the semantic context and the data features, especially for indicating trends.
% Therefore, we employed unconditional background generation for bar charts, line charts, and scatter charts.
In unconditional background generation, the aim is to incorporate semantic context without extracting objects. To achieve this, we employ a straightforward text-to-image generation by the fundamental diffusion model.
% backgroud就不需要
% cue fig3

\subsubsection{Conditional Generation}
% 总说 what，goal
Compared with unconditional generation, conditional generation involves integrating chart image $I_{c}$ to make the generated visual element comply with the data information, such as trend and contour. 
% 总说 how 解决两个chanllenges
There are two principal challenges that require addressing\cite{frans2022clipdraw,iluz2023word}:
1) Enhancing generational diversity.\red{
As illustrated in Fig.~\ref{fig:method}, the same line chart can exhibit various types of masks. These masks are provided to enhance result diversity and cater to user customization.}
We also introduce an augmentation module to expand the possible fusion directions.
\red{This allows for a wider range of design options and the iteration of the design process.}
Nevertheless, we have discovered that conventional augmentation operations used in natural image domains, such as cropping and flipping, are inappropriate for charts and may ultimately jeopardize the data integrity of the chart\cite{li2022structure}.
2) Integrating the semantic context and the chart.
This entails determine how to condition the generation process by merging the attention map containing semantic context and chart with the data information.
% 1. augment the chart mask， 增加生成多样性，和natural iamge不一样，我们要在不破坏数据完整性的基础上对image增强。 
% 2. reshape the Iobject to fit the Ic. Align the pixel // SharpContour

\noindent \textbf{Foreground.}
% There are three types of charts that require conditional generation for embedding the type of foreground internal single.
The conditional foreground generation emphasizes the integration of semantic context into the visual marks in the chart, while adhering to the data represented within the chart.
Intuitively, the semantic context needs to be integrated into the rectangle, line, sector, and bubble for the bar chart, line chart, pie chart, and scatter plot, respectively.
Initially, we randomly augment $I_{c}$ with various manipulations, including Gaussian blur, dynamic blur, and image warp, as depicted in Fig.~\ref{fig:method} (a).
The augmentation module $aug$ is established based on the principle of enhancing the diversity of chart element shapes while maintaining data integrity.
Then, we obtain the attention map $A$ concerning the $P_{object}$ from the generation process (Eq. \ref{eq1}).
To infuse the data information from $I_{c}$ into the attention map, we utilize $I_{c}$ as a mask, ensuring the attention map possesses the same shape as the element in $I_{c}$.
To maximize the fused image $I_{fuse}$ by including as much semantic context as possible, we employ two common affine transformations, scaling and rotating.
The optimization function can be expressed as:
\begin{equation}
\label{eq5}
\vspace{-0.5em}
I_{fuse} = \max_{\theta, s}
 [aug(I_{c}) \odot \phi (f_{upsample}(A),\theta ,s)],
\end{equation}
% where $s$ and $\theta$ denote the parameters for scale and rotate transformations, respectively, and $\phi$ symbolizes these transformation functions.
where $\phi$ is the affine transformation parameterized by scaling parameter $s$ and rotation parameter $\theta$.
Finally, we take $I_{fuse}$ as input, which integrates semantic context and chart information, yielding $I_{g}$.
% \begin{equation}
% \label{eq6}
% I_{g} = G[A \odot aug(I_{c})]
% \end{equation}

% We summarize conditional foreground generation as a five-step process.
% (1) Firstly, we conduct an initial generation to get $I_{g}$ containing the semantic context.
% (2) Then we extract the object based on the cross-attention between $P_{object}$ and $I_{g}$, yielding $I_{object},$ which is similar to the method used in foreground unconditional generation.
% (3) Next, we randomly augment $I_{c}$ with predetermined manipulations, such as Gaussian blur, dynamic blur, and image warp, as shown in the figure.
% The criteria for selecting the augmentation is that it must distort the chart element while preserving the data integrity.
% (4) During the fusion process, we reshape $I_{object}$ to align with the augmented $I_{c}$ in a pixel-wise manner.
% (5) Finally, we encode the fused image injected both semantic context and chart information, and regenerate to get the $I_{g}'$.
% foreground
% 总说 how 3-step
% 1. 锁定目标根据p object => I object. similar xxx
% 2. augment the chart
% 3. reshape the Iobject to fit the Ic. Align the pixel
\begin{figure}[t]
    \centering
\includegraphics[width=\linewidth]{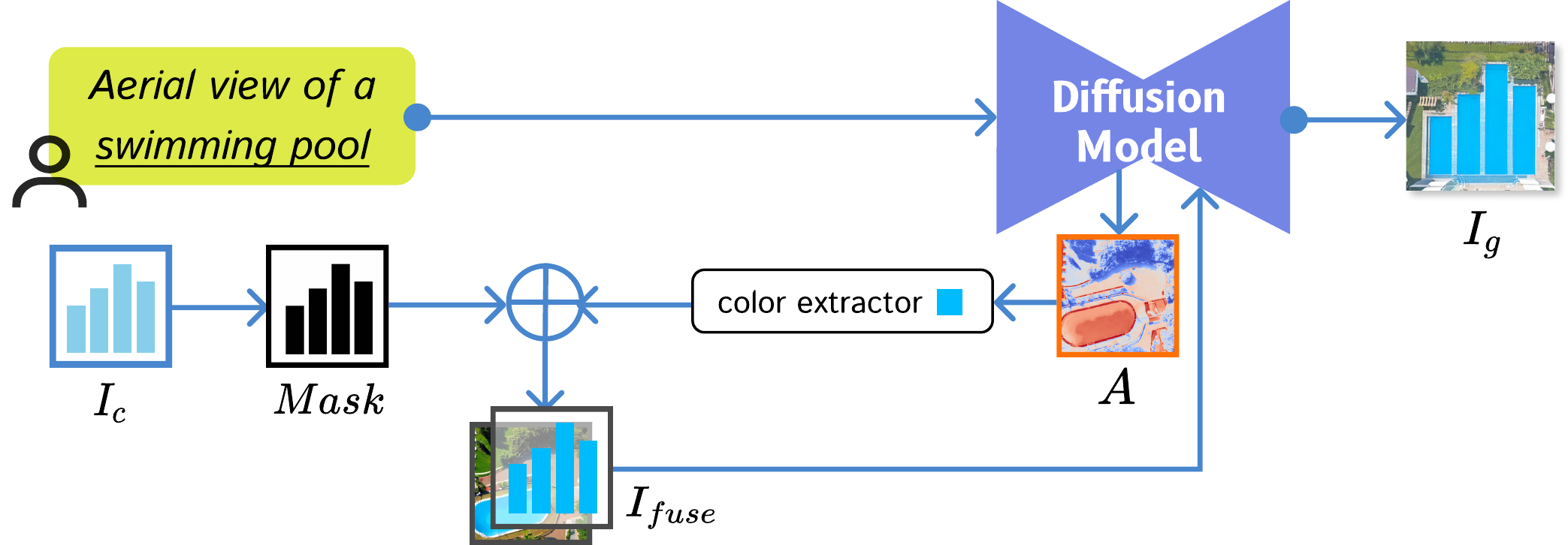}
    \caption{\textbf{The Conditional Generation for background.} We employ the color extractor based on the attention mechanism to enhance the fusion between chart and semantic context.}
    \label{fig: Con_B_g}
     \vspace{-1.4em}
\end{figure}
\noindent \textbf{Background.}
% background
% 2. augment the chart
% 3. reshape the Iobject to fit the Ic. Align the pixel
% Based on our preliminary study, we have found that the most common usage of embedding semantic context into a background is adopted by line charts.
The background serves not only as a container of semantic context but also as a part of the chart that conveys data information.\red{
To ensure seamless integration between the chart's features and the background component, we devise a novel fuse strategy incorporating color and context semantics.
In practice, we extract the dominant color based on the attention map, and then synthesize a color mask with the injection of $I_{c}$.
The resulting color mask is concatenated with the semantic information to obtain $I_{fuse}$.
Then we injected the $I_{fuse}$ into the generation procedure to achieve reconstruction, yielding the $I_{g}$.}

\subsubsection{Modification}
\label{subsub:extention and refinement}
% 总说
To improve the harmony and consistency of generated results, we devise two modification methods.
At the element level, we reuse the generated elements to encode other visual marks in the chart, enhancing its reproducibility and adaptability.
At the chart levels, we refine the image details to ensure a cohesive overall design, particularly when merging independently generated visual elements.
\red{The main technology we use for these modifications lies in image-to-image generation, which injects the initial image as a condition and we take it to improve the overall consistency of the output image.}
\begin{figure}[t]
    \centering
\includegraphics[width=\linewidth]{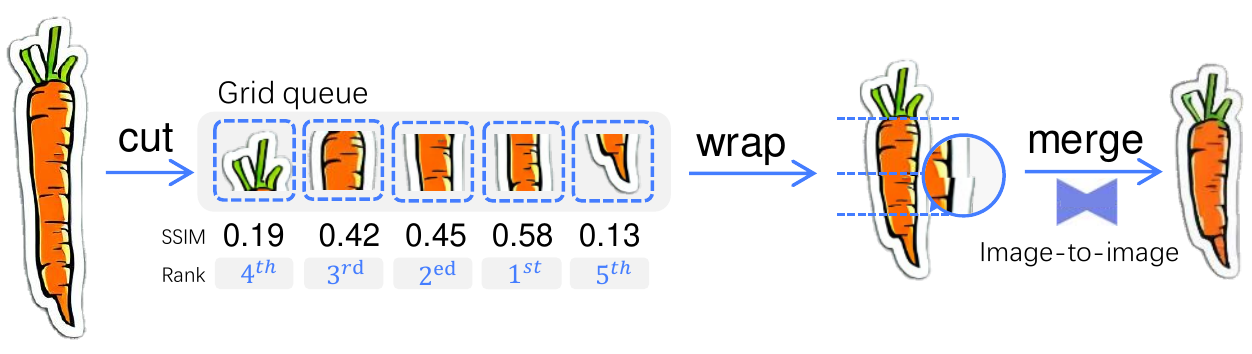}
     \vspace{-1.8em}
    \caption{\textbf{The Replication process.} It features three main steps including cut, wrap, and merge, which aim to replicate the origin visual element while changing its scale such as height.}
    \label{fig:replication}
     \vspace{-1.4em}
\end{figure}

\noindent \textbf{Replication.}
To apply the visual element to other visual marks in a chart, traditional tools require copying each individual element which is tedious and can cause distortion. 
We propose a warp-and-merge strategy shown in Fig.~\ref{fig:replication} to overcome these challenges, especially for bar charts.
We start by using the tallest bar in the chart as a reference point to generate the fundamental visual element. \red{This simplifies the task of adapting the element to the shorter bars in the chart.}
We cut the visual element into five equally elevated grids and compute the structural similarity (SSIM) amid each pair, \red{with higher SSIM indicating greater editability. 
When wrapping the grids, we scale their height to match the new bar.
However, this can result in artifacts and rigid junctions, which are addressed by using a generative model's image-to-image pipeline to merge the grids seamlessly. This also helps to make each visual mark unique instead of simply replicating them.}
% 1，按照最长的生成， 所以接下是朝着减去重复度最高的部分做优化
% 2. 分成几等分，并计算两两之间的SSIM结构相似度
% 3. 按照bar的比例，cut掉相似度最高的部分，并把剩下的拼接在一起
% 4. 由于拼接后的图像会有伪影和僵硬的衔接 将继续refine优化这些细节

\noindent \textbf{Refinement.}
During the generation process, users may integrate multiple embedding methods, yielding several independent generation results.
Directly concatenating them can give rise to incoherent styles, as depicted one the left side of the canvas of Fig.~\ref{fig:interface}.
To solve this, we also take an image-to-image pipeline to harmonize the style of the image while preserving its layout and semantic context.
The refined examples in the Fig.~\ref{fig:interface} which can be triggered by button \raisebox{-.2\height}{\includegraphics[width=0.32cm]{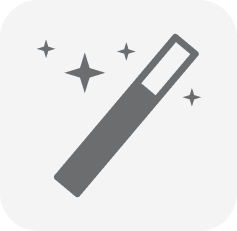}}.

% For instance, as depicted in Fig.~\ref{fig:interface}, the tree branch and the cherry blossoms are generated separately, employing unconditional and conditional foreground modes, respectively.
% In the process of refinement, we also offer user with the option to input the prompt to 
% 由于在生成过程中，用户可能会混合多种方式，举例子，用FIS的方式单独生成树枝，用FE的方式生成花朵。
% 单独生成的元素简单拼接会有几个问题：
%% 1. 拼接很不自然 collage
%% 2. 风格也不一致
% 将元素放置好在经过一次img2img的生成，可以使得风格更加统一，且凭借生成模型强大的构图能力补充细节。同时preserve location and semantic context
% 并且用户可以在这个过程中，输入其他的prompt如指定风格等等，promotexxx
\subsection{Evaluation}
\begin{figure*}[t]
    \centering
    \includegraphics[width=0.98\linewidth]{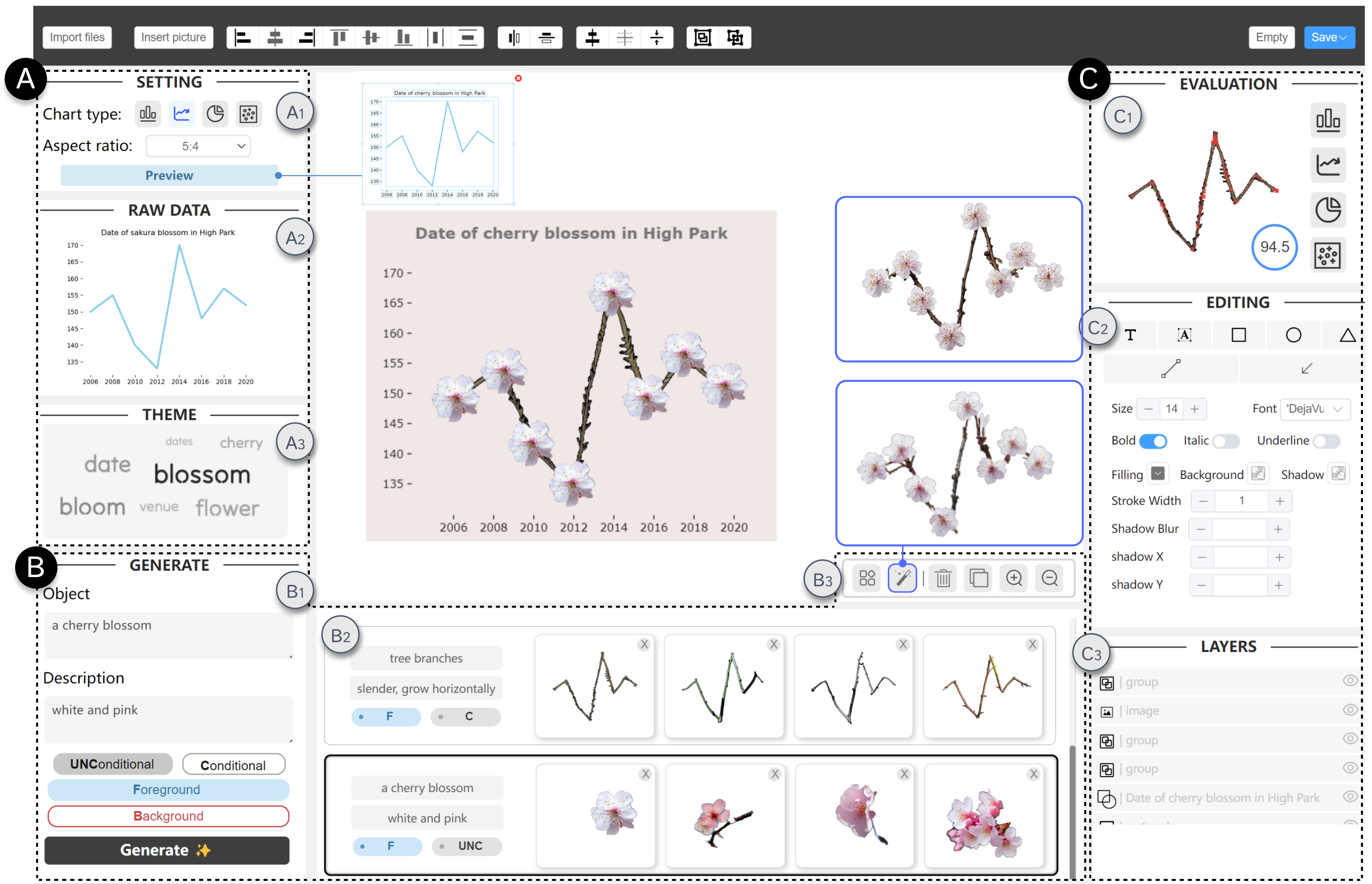}
    \caption{\textbf{User interface of \textit{ChartSpark}.} It consists of a central canvas for the manipulation and composition of visual elements and three main modules corresponding to the process of feature extraction (A), generation (B), as well as evaluation and editing (C).}
    \label{fig:interface}
    \vspace{-2em}
\end{figure*}

\red{Generation in data visualization can raise concerns about compromising the integrity of the data and collapsing as ``\textit{chartjunk}''\cite{tufte1985visual,few2011chartjunk}.}
As stated in requirement \textbf{R4}, it is essential to provide an evaluation module to inform users of any potential distortions that affect data integrity.
\red{There are two ways to achieve it:
1) The plain chart with each element editable in canvas (Fig.~\ref{fig:interface}) can be used to compare with the generated results including conditional and unconditional.
2) A evaluation module displays global distortion value and local regions with errors (Fig.~\ref{fig:interface} ($C_1$)).
We elaborate on the latter method in the following graph.}
% In order to ensure a faithful representation of the data, we provide a data distortion assessment to inform users of any potential distortions.

% \subsubsection{Distortion}
% 从那几个角度

% To assess distortion, we ascertain the disparity between the generated visual element and the original plain chart (Fig.~\ref{fig:interface}($C_1$)).
The evaluation module is designed for conditional generation which considers data in its generation process.
Given that each chart employs distinct visual channels to encode data, we tailor our methodology to guarantee reliable evaluation for each chart type.
For bar charts, we concentrate primarily on height as an indicator of distortion.
For line charts, the portrayal of the trend is of paramount importance.
For pie charts, we measure the angle for each sector.
For scatter charts, we estimate the size of each point.
% bar chart => height
% line chart => trend -- 中轴线
% pie chart => angle
% scatter chart => 质心
Compared with the approach in \cite{coelho2020infomages}, \textit{ChartSpark} not only provides numerical values but also displays locations with high errors to facilitate modification.

% \begin{figure}[h]
%     \centering
% \includegraphics[width=\linewidth]{figs/evaluation_line.pdf}
%     \caption{\textbf{Evaluation process.} For line chart (a) and generated element (b), we go through the 3 steps: extracting the line/central axis, taking box filter, and sliding by the window. (c) is the generated element with red error indicators.}
%     \label{fig:evaluation_line}
%      \vspace{-0.4em}
% \end{figure}

\begin{figure}[t]
    \centering
\includegraphics[width=\linewidth]{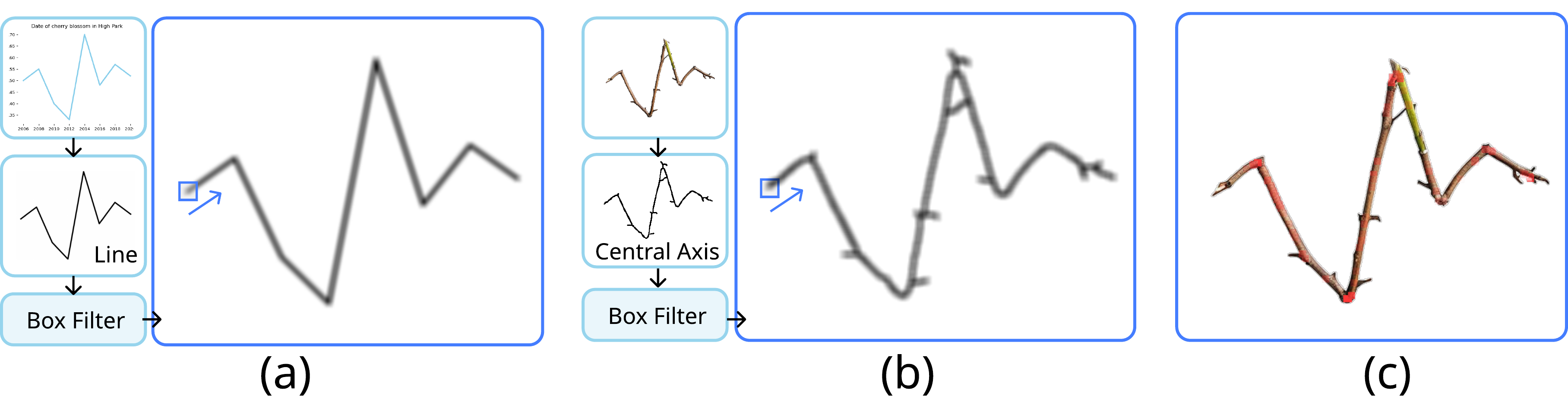}
    \caption{\textbf{Evaluation process.} For line chart (a) and generated element (b), we go through the 3 steps: extracting the line or central axis, taking box filter, and sliding by the window. (c) is the generated element with red error indicators.}
    \label{fig:evaluation_line}
     \vspace{-1.4em}
\end{figure}

\noindent\textbf{Foreground.} 
It is relatively simple to compute the distortion for bar, pie, and scatter charts, as we can directly calculate the metrics and then compare them with the origin data. 
However, it's not straightforward to compare the trend between generated visual element and the line chart (Fig. \ref{fig:evaluation_line}). Unlike Infomages\cite{coelho2020infomages}, which focus on the line slope, we take the evaluation in pixel-wise.
We extract the central axis from the generated element, smooth the pixel values with a box filter, and use a window to slide along the chart $I_{c}$ and generated element$I_{g}$.
In each window, we will compute the mean pixel values to get the score as follows:
\begin{equation}
\label{eq7}
\vspace{-0.5em}
S_{i} =  1- \frac{\left | W^{i}_{I_{c}} - W^{i}_{I_{g}}\right |}{255},
% \vspace{-0.3em}
\end{equation}
The higher the score, the greater the similarity between the two windows.
Additionally, We visualize the window of this score, with a red square representing the distortion region.
We get the global score by averaging each window score along the line.

\noindent\textbf{Background.}
To simplify the evaluation process, we utilize edge detection techniques that effectively identify prominent boundaries and edges within the image.
It allows us to eliminate any redundant and distracting details, enabling us to focus on capturing the precise outline of the target object.
Furthermore, to improve the saliency of data-related semantic objects in the image, we use the chart mask to filter out irrelevant objects.
Now the problem is simplified and can be approached similarly to the foreground evaluation.

% \begin{equation}
% \label{eq8}
% \vspace{-0.5em}
% S = \frac{1}{N} \sum_{i=0}^{N} S_{i},
% \end{equation}
% where the $N$ represents the total number of the window that will go through.

% Firstly, we extract the central axis from the generated visual element, which can approximately represent the data.
% Then we take the box filter with the size of 7*7 for both the central axis and the line chart, aiming to smooth the value between pixels.
% Then we leverage a window with the size of 7*7 along the data in the line chart with a stride of 5.
% We take the window in the according location in the central axis.
% Through the pixels in the window, we can calculate the mean values Vc and Vg for chart and generated visual element, respectively.
% The score can be represented as xxx.
% The higher the score, the more similar the generated visual element and chart are.
% Additionally, we visualize each window using its score, with the conspicuous red square representing the distortion region.

% \subsubsection{Disharmony}
% 元素颜色和谐程度

% 前后景遮掩程度（有些 独立生成的）overlapping
\begin{figure*}[t]
    \centering
    \includegraphics[width=\linewidth]{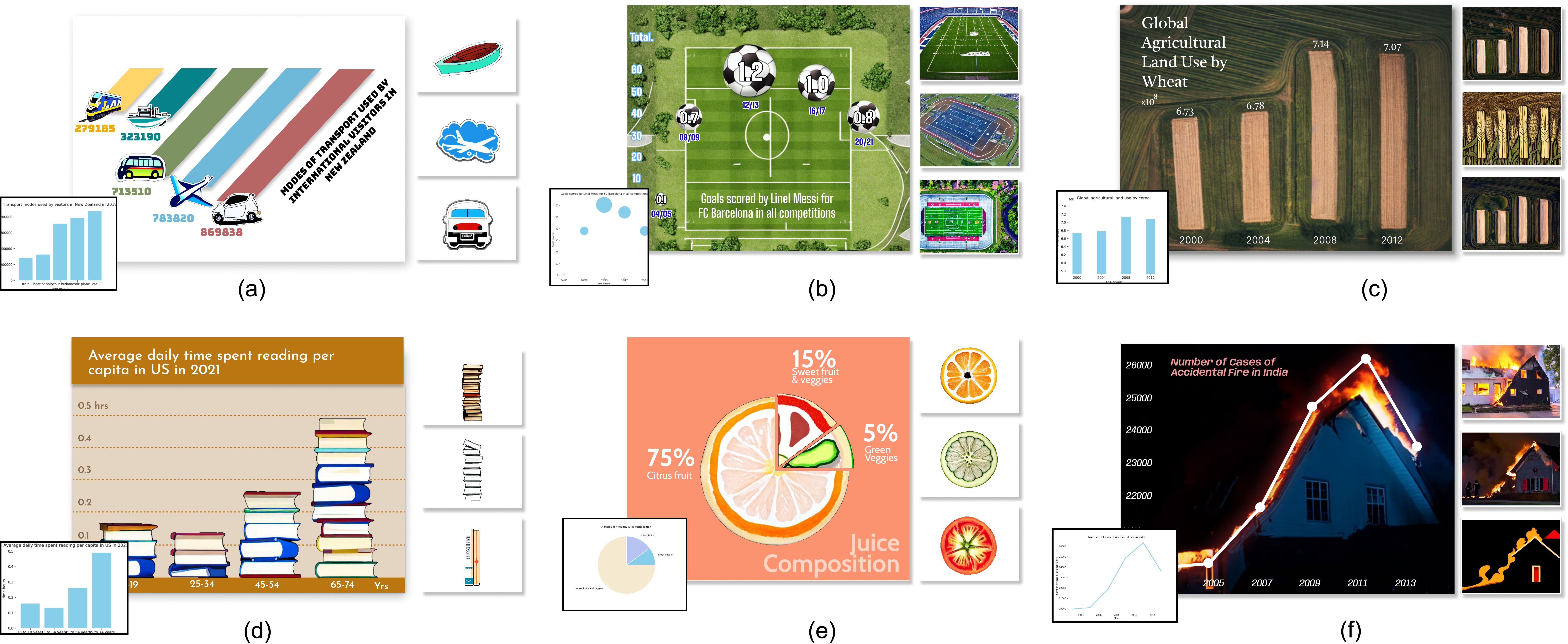}
    \caption{\textbf{Pictorial visualizations generated by \textit{ChartSpark}}. (a) A bar chart showing the transportation types used by visitors in New Zealand. (b) A scatter plot showing Messi's total goal scored and average goal scored for FC Barcelona per season. (c) A bar chart showing the global agricultural land used by cereal. (d) A bar chart showing the average daily reading time per capita in US in 2021. (e) A pie chart showing a juice composition recipe. (f) A line chart showing the number of Indian fire accidents per year.}
    \label{fig:case}
    \vspace{-1.7em}
\end{figure*}

%% file: Latex/5_Interface.tex
% \begin{figure*}[t]
%     \centering
%     \includegraphics[width=0.95\linewidth]{figs/case.pdf}
%     \caption{\textbf{Pictporial visualizations generated by \textit{ChartSpark}}. (a) A bar chart of the transportation types used by visitors in New Zealand. (b) A scatter plot of Messi's total goal scored and average goal scored for FC Barcelona per season. (c) A bar chart of the global agricultural land used by cereal. (d) A bar chart of the average daily reading time per capita in US in 2021. (e) A pie chart of a juice composition recipe. (f) A line chart of the number of Indian fire accidents per year.}
%     \label{fig:case}
% \end{figure*}
\section{Interface}
The user interface (Fig.~\ref{fig:interface}) is divided into three core modules that correspond to the process of feature extraction ($A$), generation ($B$), and evaluation and editing ($C$).
The central canvas of the interface allows users to flexibly manipulate elements\red{, such as rotating and scaling}.

\noindent \textbf{Feature extraction.}
To begin, the user uploads a data file and selects a desired chart type and aspect ratio in the settings panel ($A_1$).
The raw data is displayed in $A_2$, and the relevant semantic context is shown in $A_3$.
The data annotation information will also be rendered on the canvas.
Both the visual representation of the raw data and the semantic context aid users in understanding the data and supplying a design precedent before creation.
They are displayed throughout the creation process, enabling users to compare the generated visualizations with the original data (\textbf{R1}).

\noindent \textbf{Generation.}
The Generation module comprises three components: generation options ($B_1$), gallery ($B_2$), and modification ($B_3$).
In $B_1$, users can customize the generation object and generation styles utilizing text boxes and select the generation target and method through two sets of buttons (\textbf{R2, R3}). The generation results are displayed in $B_2$, labeled with the generation options. 
Users can choose whether to preserve or discard the generation element in the $B_2$, and regeneration can be performed from $B_1$. The $B_2$ view is closely linked to $B_1$, as the content \red{in the row user clicks in $B_2$ will accordingly be shown in $B_1$, promoting the backtrack of generation.} The first two buttons of $B_3$ serve as modifications in the generation process, corresponding to the replication and refinement functions described in Sect. \ref{subsub:extention and refinement}.

\noindent \textbf{Evaluation and editing.}
$C_{1}$ offers users data distortion evaluation with both explicit qualitative value and error visualization (\textbf{R4}).
$C_{2}$ comprises several tools allowing users to edit their visualizations, including font (such as type, size, and bolding), basic shapes, effects, and strokes.
$C_{3}$ displays the layers of elements, and provides the ability to adjust the layer order and visibility.

% transform (such as scaling, rotating)
% In section B, users can selected a generation method to generate visual elements according to their input object and descriptions.
% Users have the option to keep or discard the generated results.
% Multiple generation are allowed by the same or different objects, descriptions and generation methods.
% Section C provides users with the ability to comprehend potential distortions and disharmonies in their visualizations through estimation values and corresponding thermodynamic diagrams.
% Section D allows users to edit their visualizations, including images (such as scaling, rotating, and translating) and text (such as font, size, and bolding).
% Users can also adjust the layers for their visual elements to facilitate editing.

%% file: Latex/6_Evaluation.tex
\section{Evaluation}
In order to comprehensively evaluate the efficacy of \textit{ChartSpark}, we performed three distinct analyses, including example applications, a user study, and an expert interview.

\subsection{Example Applications}

Based on the taxonomy used in the preliminary study, we utilized \textit{ChartSpark} to generate the four primary categories of pictorial visualizations.
As illustrated in Fig.~\ref{fig:case}, we present the basic chart for each pictograph in the lower left corner, while showcasing the alternative visual elements generated on the right side.

\noindent
\textbf{Foreground External.}
As shown in Fig.~\ref{fig:case} (a), the shape of visual elements in this pictorial visualization is not constrained by the data.
Instead, they function as embellishments to the chart.
% 生成方式
To generate these elements, we have adopted the unconditional foreground mode, which enables us to extract the target through $P_{obj}$, while using the same $P_{des}$, which we have defined as \textit{``a well-designed sticker''}, ensures consistency in the style of different transportation representations. 
% 系统editing module里面
% Additionally, the editing modules in \textit{ChartSpark} offer several tools to enhance the visual effect.

\noindent
\textbf{Foreground Internal Single.}
Fig.~\ref{fig:case} (b) and (e) show using the generated visual element to replace the original visual mark.
% in a more a visually appealing way. 
In (b), the generated pictograph employs the same encoding as the original chart, with height signifying goals scored during a specific season and size indicating the average goal.
Football elements replace the bubbles, and a football pitch is generated to complement the foreground.
Fig.~\ref{fig:case} (e) displays the cross-section of each fruit and vegetable supplanting the original fan-shaped segments in the pie chart.

\noindent
\textbf{Foreground Internal Multiple.}
Fig.~\ref{fig:case} (d) involves multiple replication units within a single bar. 
% Traditional tools present two challenges in this process. 
% two challenges
% Firstly, users may need to replicate a single book element several times to fill the bar, leading to a monotonous and unnatural visual effect. Secondly, the different heights of each bar can impede the reusability of created elements, and scaling cuts may compromise the final effect.
\textit{ChartSpark} provides various visually diverse elements that fit the height of the bar using the same prompt, \textit{``the pile of books''}. Each book in the element is unique, circumventing monotonous duplication.
Moreover, each bar is filled with the element naturally and automatically without manual manipulation.
% Additionally, users can automatically apply the visual element to other bars using the replication (add icon) while maintaining its style.

\noindent
\textbf{Background.}
In Fig.~\ref{fig:case} (c) and (f), the chart has been merged into the image in a more natural manner under the mode of \textit{background} and \textit{conditional}, which \red{encodes both} the semantic and data information.
In case (c), as the data describes the land use of wheat, the prompt is utilized as \textit{``aerial view of wheat field''}.
In case (f), the fire trend of the generated image aligns with the original chart.

\subsection{User Study}
We conducted a user study to evaluate the usability and effectiveness of our tool to facilitate creation of pictorial visualization.
The study consists of a survey, succeeded by a semi-structured interview to obtain more in-depth qualitative feedback.

% about our method and tool.

\noindent
\textbf{Participants}:
We recruited 8 participants aged between 21 to 28, who are visualization users interested in pictorial visualization.
The participants are graduate students majoring in different disciplines from computer science, business, art and design, to architecture.
All participants have used or designed pictorial visualization.

\noindent
\textbf{Procedure}:
The study was conducted in a one-on-one and face-to-face manner.
First, we showed some examples of pictorial visualization and introduced to the participants the basic concepts including the embedding objects (foreground and background) and embedding techniques (conditional and unconditional).
% After that, we briefly explain the previous tools for creating pictorial visualization and the goal and design considerations of our authoring tool.
Next, we presented the interface to the participant and introduced the functionalities.
We then guided the participant through a step-by-step process of creating a pictorial visualization. Finally, we encouraged them to independently explore the authoring tool for 10 minutes to gain further familiarity.
% During the exploration period, we answer their questions about the functions
Afterwards, participants were asked to make a pictorial visualization from given data files.
We carefully observed their creation process and documented their inquiries and remarks.
Upon finishing the creation, the participant took the survey and answered four 5-point Likert scale questions.
The survey was followed by a short interview for additional feedback.
Each study lasted for about 50 minutes.
Participant were compensated with a gift of 10\$ after completing the study.

\begin{figure}[t]
    \centering
    \includegraphics[width=\linewidth]{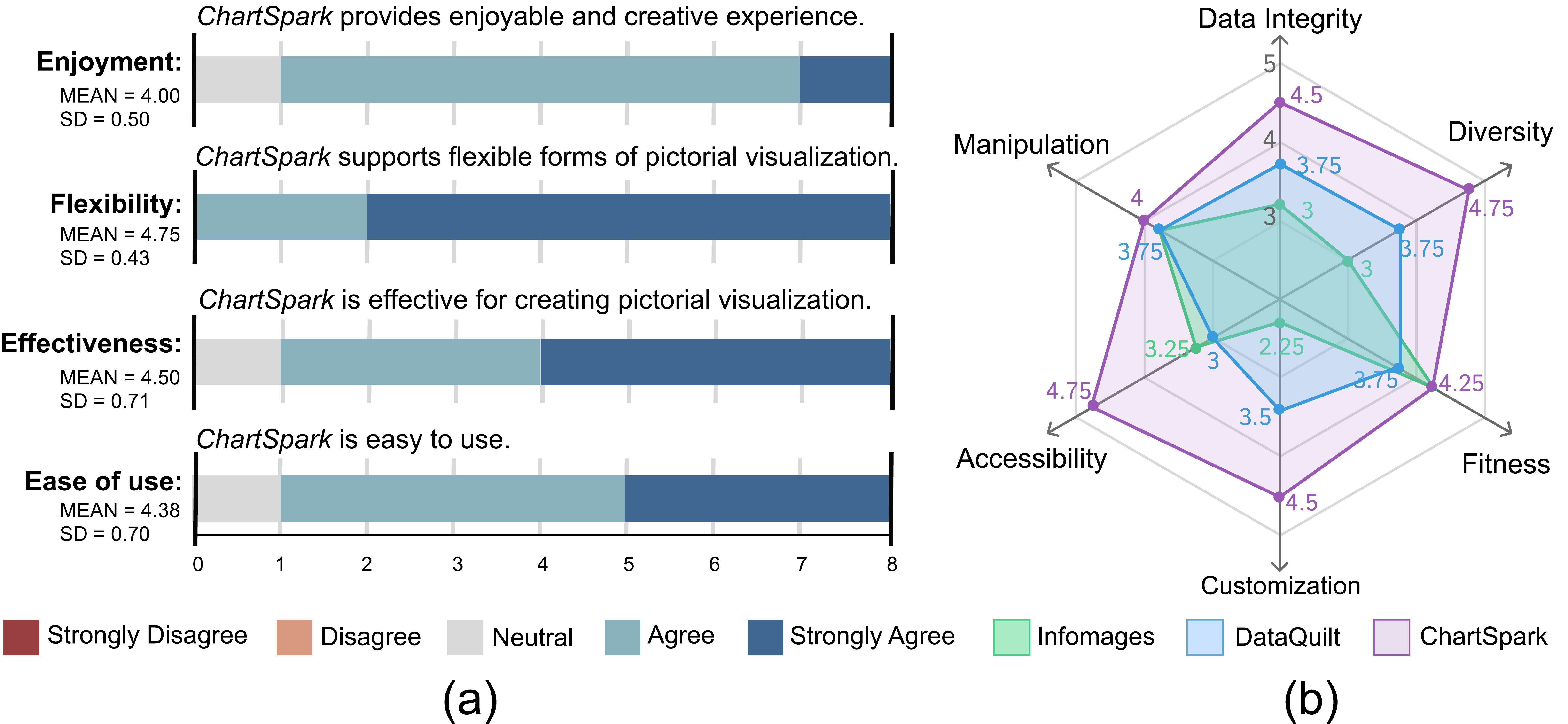}
    \caption{\textbf{Results of user study and expert interview}. (a) represents the result of users' evaluations in enjoyment, flexibility, effectiveness, and ease of use, with mean and standard deviation respectively. (b) is \red{the comparison of} \textit{ChartSpark} with DataQuilt and Infomages.}
    \label{fig:expert}
    \vspace{-1.6em}
\end{figure}

\noindent
\textbf{Result Analysis}:
% 数据
%% 
%% 
We report our survey results here, see Fig.~\ref{fig:expert} (a).
% \begin{figure}
%     \centering
%     \includegraphics[width=0.85\linewidth]{figs/User_study/User_Results.pdf}
%     \caption{\textbf{Result of user study}. The proportion of participants' evaluations in enjoyment, flexibility, effectiveness, and ease of use, with mean and standard deviation respectively.}
%     \label{fig:user}
%     \vspace{-2em}
% \end{figure}
% \begin{itemize}
% \setlength\itemsep{-0.28em}
\begin{itemize}[leftmargin=*,topsep=0.25em]
\setlength\itemsep{-0.28em}
\item
\textbf{Enjoyment}:
Notably, the novel creative experience provided by our tool proved to be delightful for the majority of participants ($mean=4.00, \; SD=0.50$).
P1 stated, \emph{``I really enjoy engaging with the AI and see all the interesting possibilities generated by the tool.''}
\emph{``I have tried many popular text-to-image tools, this is the first that enables me to employ AI in visualization. It has heightened my engagement in the creative process,''} commented P8.
\item
\textbf{Flexibility}:
The majority of participants praised the flexibility of our tool ($mean=4.75, \; SD=0.43$).
P3 said, \emph{``I could freely switch between foreground elements and background elements and combine them to create a variety of interesting pictorial visualizations.}''
P7 added, \emph{``This tool driven by AI generative model helped me find design materials of diverse styles which are not easily available in traditional design pipeline where I had to spend much time searching for various references.''}
\item
\textbf{Effectiveness}:
All participants acknowledged the effectiveness of our system for creating pictorial visualization ($mean=4.50, \; SD=0.71$).
\emph{``As I am not adept at professional graphic design tools like Adobe Illustrator, this tool makes it possible to many people like me,''} P6 remarked.
\textit{``I'm amazed that the generated elements can comply with the data trend, and there is a wide variety of options.''}
\item
\textbf{Ease of use}:
Most participants agreed that the authoring tool is easy to use ($mean=4.38,\; SD=0.70$).
\emph{``Most of the functions are intuitive, easy to understand and use,''} P5 commented.
\emph{``The whole creation process is smooth because of the simple design of the interface and the convenient functions supported by AI such as the replication button,''} P1 said.
\emph{``However, I think the object and description can be put into one text entry box to make it more convenient,''} P2 suggested.
\end{itemize}

% 采访
\noindent
\textbf{Feedback}:
We summarize the feedback from the interview as follows.
\begin{itemize}[leftmargin=*,topsep=0.25em]
\setlength\itemsep{-0.28em}
\item
\textbf{One-shot generation vs. more intermediate control}:
Some novice users may prefer quick generation without much human intervention.
For example, P4 said, \emph{``It would be really efficient if I only need to import the data and the tool just generate ready-made pictorial visualization for me.''}
However, users who are more experienced in design favor more control of the intermediate process and the results.
\emph{``Even though AI can provide impressive inspiration, it is better for me if I can still pick what materials I want to use like in this tool. But I guess more control such as by sketch would also be beneficial.''}
\item
\textbf{Faithful evaluation}:
The users also talked about \textit{CharSpark} supporting reliable data presentation.
P3 said, \emph{``I worry that pictorial visualizations may overly emphasize aesthetics, sacrificing readability. Thus, I value the design featuring editable data annotations, saving me the trouble of adding sticks and labels manually.'' P5 mentioned, \emph{``The evaluation view displays precise error location, allowing me to easily identify problematic areas, which is far more intuitive than simply presenting numerical values.''}}
% \item
% \textbf{Gallery-based authoring tool}:
% The users also talked about the benefits of a gallery-based authoring tool.
% P3 said, \emph{``I think the gallery is a very useful design for authoring tool because I can directly see the references during my creation. Combined with AI's generation power to quickly update the content in the gallery, this function is further enhanced.''}
\item
\textbf{Integrated tool}:
Users also pointed out the importance of an integrated tool.
\emph{``Nowadays many AI applications do not come with post-editing functions. As a result, I often have to export the AI-generated result to external tool like Photoshop for editing. So I appreciate the effort to add several editing functions, but I also suggest adding more advanced functions like those in Photoshop or develop a plugin for Photoshop,''} P1 commented.
\end{itemize}

\subsection{Expert Interview}
To conduct a thorough comparison of the performance of \textit{ChartSpark} with previous tools, including DataQuilt \cite{zhang2020dataquilt} and Infomages \cite{coelho2020infomages}, we engaged four designers and experts with extensive experience in information design to evaluate these tools.
Designer 1 (D1) has expertise in graphic design and has been working in the information design field across both print and digital media for over six years.
Designer 2 (D2) specializes in UX/UI design and has been working in this field for over three years.
Designer 3 (D3) focuses on data storytelling and interactive art, and is familiar with AI-driven techniques for creating engaging visual content.
Expert 1 (E1) has been engaged in both academic and industry projects for over four years, focusing on the development and evaluation of visualization techniques and tools.

We established six metrics that comprehensively cover the creation of pictorial visualizations for detailed assessment of each tool.
\textbf{Data Integrity} evaluates the system's response to data errors and the tool's ability to communicate the errors to users.
\textbf{Generalization} examines the tool's ability to support a broad range of methods to incorporate semantic context into charts.
\textbf{Fitness} reflects the successful integration of visual elements and charts, taking into account aspects such as trend.
\textbf{Customization} evaluates the tool's capacity to tailor visual elements and offer a diverse set of customization options.
\textbf{Accessibility} depicts the ease with which visual elements can be obtained for design purposes.
\textbf{Manipulation} measures how effective the tool is to help users alter visual elements for the specific needs of charts.
% \begin{itemize}[leftmargin=*]
% \setlength\itemsep{-0.28em}
%     \item \textbf{Data Integrity.} This metric evaluates the system's response to data errors and the tool's ability to communicate the errors to users.
%     \item \textbf{Generalization.} This metric examines the tool's ability to support a broad range of methods to incorporate semantic context into charts.
%    \item \textbf{Fitness.} This metric concentrates on the successful integration of visual elements and charts, taking into account aspects such as trend.
%    \item\textbf{Customization.} This metric evaluates the tool's capacity to tailor visual elements and offer a diverse set of customization options.
%    \item \textbf{Accessibility.} This metric focuses on the ease with which visual elements can be obtained for design purposes.
%    \item \textbf{Manipulation.} This metric measures how effective the tool is to help users alter visual elements for the specific needs of charts.
% \end{itemize}
The first metric focused on faithfulness, while the second to fourth metrics focused on the expressiveness of pictorial visualization. The last two metrics evaluated the usability of the tools. 
% 实验流程
To begin the evaluation process, we presented several cases to each participant, revealing embedding techniques that merge semantic context and data information in pictorial visualizations. 
% video --- 全貌，整体，操作流程
Next, we showed a video of each tool, demonstrating the overall operation process and main function modules in a randomized order.
% slides --- case， function
We then presented additional cases to the participants and asked them how to create a pictorial visualization if they use these tools following the think-aloud protocol. 
% 询问细节问题 讨论
% 评分
After completing the evaluation process, each participant provided a score for each tool based on the six evaluation metrics. The results are shown in Fig.~\ref{fig:expert} (b).
%% ========================== 补充对图片的描述

\noindent \textbf{Faithfulness.}
D1 mentioned that \emph{``It is challenging to check the accuracy of data binding in DataQuilt as there is no evaluation offered.''} On the other hand, D3 pointed out that \emph{``ChartSpark offers both numerical evaluation results and shows the error location, which can guide users on how to improve the visualization.''}

\noindent \textbf{Expressiveness.}
% generalization
Regarding expressiveness, all participants agreed that \textit{ChartSpark} is more versatile in embedding semantic context to both foreground and background, whereas Infomages focus on the background and DataQuilt on the foreground.
D2 commented that \textit{``Infomages use an image as the main component, which makes customizability weak and compatibility with chart types limited.''}
% fitness
% 很喜欢用背景去表征，但是这也是最难的，比在前景中编辑更加困难
D1 appreciated Infomages' design and found that the merge of elements and charts seems very natural. However, she also mentioned that \textit{``it is difficult to find an image with the same trend as the plain chart.''}
D4 commented on the embedding methods of DataQuilt and 
Infomages, describing them as \textit{``unidirectional and full of compromises''}. They explained that \textit{``DataQuilt extracts visual elements from images and adapts them to fit the height or area of the original chart, while Infomages alters the chart's shape to accommodate the semantic image.''} In contrast, \textit{ChartSpark}'s generation results represent a bilateral merge, reshaping the visual elements intrinsically.
% customization
D3 said that \textit{``There is no much room for me to change the shape or color in elements in Infomages and DataQuilt.''}
D2 praised \textit{ChartSpark}'s customization capabilities, noting that \textit{``If I want to change the number or color of visual elements, I can achieve it by modifying the description prompt. I find the alternative element in the gallery is important for design iteration.''}
But D2 worried that the generated element may not meet the high resolution requirement. If so, she will prefer to upload the visual element.

\noindent \textbf{Usability.}
All participants mentioned the \textit{``difficulty to retrieve the element containing the exact same encoding as the given data, particularly for Infomages''}.
Especially the overlay performance of Infomages \textit{``heavily relies on the retrieved image''}.
D4 expressed appreciation for \textit{ChartSpark}: \textit{``It eliminates the need to search for images and allows to generate as many visualizations as desired.''}
However, D1 raised concerns about potential copyright issues arising from AI-generated content.
Regarding the manipulation of these tools, D2 said, \textit{``Infomages requires the least manual intervention among the tools but meanwhile \red{limits} the combination ways of \red{merging} the visual element and chart. DataQuilt primarily automates batch operations for mapping elements to data, but users still need to consider how to adapt the visualizations. \textit{ChartSpark}, on the other hand, incorporates data consideration when providing visual elements.''}
Furthermore, D3 pointed out that data annotation and axis automation are available in \textit{ChartSpark}, which could decrease the workload during the creation.
% For the manipulation of these tools, D2 think that "Infomages 是需要我动手操作最少的tool， but 限制了图片和图表结合的种类。DataQuilt的自动行主要是体现在批量操作map the element to data，但是具体的如何适应仍需要我去思考，ChartSpark则是在给出这个visual element的时候就让他考虑了数据。"
% D2 think that "the data annotation and axis is not automated in \textit{ChartSpark}, which will 增加创作负担。“

%% file: Latex/7_Discussion.tex
\section{Discussion}
By utilizing the generative model, we can embed both semantic context and data in crafting a pictorial visualization.
However, our system still has several limitations that should be addressed in future work.

\subsection{Limitation}
% We summarized three primary gaps between the user's expectations and the current performance of \textit{ChartSparrk}.

% taxomomy， embedding object/type等等
\noindent \textbf{Diversity of supported visualization.}
Currently, \textit{ChartSpark} only supports a limited set of chart types, including bar charts, line plots, pie charts, and scatter plots. This limitation can be problematic when users require more diverse chart types to communicate their data effectively.
% Secondly, the embedding object can beyond our defind set 

% However, there are many other variations such as stacked bar chart, timeline chart, and proportion chart, which also take important places in pictorial visualizations.
%bar chart, 等等
%% 混合

\noindent \textbf{Controllability of generative model.}
The results generated by our tool may be unexpected for users due to various factors.
The design of the prompt plays a crucial role in meeting specific requirements, but even minor changes to the prompt can have a significant impact on the generated output.
Additionally, there may be inconsistencies in style as the visual elements are generated independently.
Making adjustments for generated elements is not flexible.
For example, a user may want to modify the color of a generated object while preserving other attributes.
% In addition, the copyright of pictorial visualizations generated by our tool can be a controversial issue, as the training of generative models uses a significant amount of material obtained from the Internet.

\red{
\noindent \textbf{Ethics and responsibility of generative model.}
The generative model may face potential criticism on copyright or bias issues, as the training process digests a huge amount of data obtained from the web, which is unfiltered and imbalanced.
The accessibility of such models may also have downside as it increases the risk of deceptive content and copyright infringement.
% The generated results can exhibit bias due to the unfiltered and imbalanced training data.
Recent research\cite{schramowski2023safe} aims to suppress inappropriate generation like pornography and racially-charged content.
However, it is crucial for both creators and viewers of AI-generated content to adopt a critical mindset.}

% 数据的忠实表达
\noindent \textbf{Faithfulness in data expression.}
% The faithfulness of data expression can be affected in the process of creating pictorial visualization.
% Although we integrate chart images as a condition for conditional generation, there is an inevitable degree of visual distortion that can cause the resulting visual elements to deviate from the original data.
A trade-off exists between maintaining data integrity and crafting a visually appealing design.
\red{Accurately preserving the data is hard for the model as it's rare for the object in real life to have such a rigid and unnatural outline.}

% To ensure effective communication of information, users may need to add data annotations to enhance the readability of the visualization. To address this issue, we offer editable data annotation formats and an editing module equipped with various tools. However, this can increase the workload and impose design limitations on users, which requires manual adjustment of sticks and labels to ensure consistency with the overall style of the visualization.
% 衡量也有需要改进点

\subsection{Future Work}
\noindent \textbf{Enrich the design space.}
To enhance the practicality of ChartSpark, we should consider integrating more types of infographics. Furthermore, the embedding object can be subdivided into more granular and targeted categories for each specific type of chart to achieve better data communication and stronger visual expressiveness.
Currently, our augmentation modules utilize the same manipulation techniques for all types of charts. However, due to differences in visual encoding among chart types, a more targeted design is necessary to ensure effective data communication and stronger visual expressiveness.

\noindent \textbf{Improve controllability of generative model.}
% Improve Controllabilty of generation.
To improve the controllability of the generation process, we can incorporate different forms of prior knowledge and allow for editing and iteration of generated results. Besides prompt-based input, other inputs like sketches and semantic maps can be included to provide more flexible and diverse information. Furthermore, we can enhance the iteration process by allowing users to select satisfactory results as a baseline for future generations, instead of generating independently each time. To address the trade-off between a more natural semantic context and data integrity, we can add a slider to the interface for users to adjust the level of emphasis on each aspect to their specific needs and preferences. Additionally, we can enable users to edit the generated visual elements through an instruction-based approach~\cite{brooks2022instructpix2pix} for localized editing instead of regenerating from scratch. 
% This not only reduces the need for regenerating the entire visualization but also enables fine-grained editing for localized changes in the visual elements. 
% In addition, we should provide more diverse condition method to improves controllability by reducing the sensitivity and inconsistency of the generation process, such as 
%% 以更多形式提供generation prior， 比如sketch，semantic map等等。

\noindent \textbf{Enhance faithfulness in data expression.}
% Generation Performace Detection.
To enhance the faithfulness of \textit{ChartSpark}, there can be two areas of improvement. 
One is to further enhance the evaluation module, as numerical values and error locations may not be explicit enough to guide them in rectifying visual elements, and more detailed inferences and specific suggestions should be provided.
% Although numerical values and error locations are provided to users, it may not be explicit enough to guide them in rectifying visual elements. Therefore, more detailed inferences and specific suggestions should be provided.
The other way to improve faithfulness is to mitigate potential distortions that may occur during the creation process. 
It is important to conduct thorough detection and analysis at each stage of the generation process for identifying potential issues like unsuccessful element extraction or poor background removal. 
% This can help identify any potential issues or errors that may compromise the performance of the generation process. 
% For instance, when target objects are not appropriately identified, unsuccessful element extraction may occur, resulting in poor-quality visualizations. Similarly, background removal may perform poorly if incomplete removal is done or if the wrong part of the image is removed. 
By detection and analysis, users can get better generated images avoiding various failures that can compromise the overall quality and accuracy of the visualizations.

% \noindent \textbf{Evaluation \& Improvement}
% Currently, our evaluation module informs users the visual distortion through numerical values and thermodynamic diagrams.
% However, users may not be able to conduct effective modifications based on their limited comprehension of the given information.
% Therefore, more detailed inferences and specific suggestions should be provided to better understand and address the problems in current pictorial visualization.
% Enrich
%% chart type 只有四种， 没有考虑到其他的比如stacked bar chart, 等等

%% file: Latex/8_Conclusion.tex
\section{Conclusion}
In this paper, we propose a novel framework named \textit{ChartSpark} to create pictorial visualizations. The framework employs a text-to-image generative model to integrate both the semantic context and chart information. \textit{ChartSpark} is flexible and compatible with both unconditional and conditional methods to fuse the semantic context with charts. It is also versatile and can be applied to both foreground and background visualizations. Additionally, we have developed a user interface that integrates feature extraction, generation, and evaluation to facilitate the creation process.
To evaluate the effectiveness of \textit{ChartSpark}, we present several cases covering the four main forms of pictorial visualizations. The feedback from user studies and expert interviews demonstrates the framework's effectiveness.
Source code of the model and interface, and the pictorial visualization corpus are released at \url{https://github.com/SerendipitysX/ChartSpark} to promote future work in this direction. 